\documentclass[11pt]{article}

\usepackage[]{acl}

\usepackage{times}
\usepackage{latexsym}
\usepackage{booktabs}
\usepackage{nicematrix}
\usepackage{multirow}
\usepackage[T1]{fontenc}
\usepackage[utf8]{inputenc}
\usepackage{microtype}
\usepackage{inconsolata}
\usepackage{graphicx}
\usepackage{verbatim}
\usepackage{tabularx}
\usepackage{array}
\usepackage{xcolor}
\usepackage{bm}
\usepackage{xurl}

\usepackage[dvipsnames]{xcolor}
\usepackage[table]{xcolor}
\usepackage{tcolorbox}
\tcbuselibrary{listings,breakable,skins}

\usepackage{fancyvrb}
\usepackage{microtype}
\usepackage{listings}
\usepackage{inconsolata} 
\newcolumntype{Y}{>{\raggedright\arraybackslash}X}

\newtcolorbox{keyfinding}[1]{%
  enhanced,
  breakable,
  colback=blue!3,
  colframe=blue!45,
  boxrule=0.35pt,
  sharp corners,
  left=7pt,right=7pt,top=8pt,bottom=8pt,
  before skip=10pt, after skip=10pt,
  borderline west={2pt}{0pt}{blue!70!black},
  overlay={%
    \node[
      fill=blue!12,
      text=blue!70!black,
      rounded corners=2pt,
      inner xsep=5pt,
      inner ysep=1.5pt,
      font=\bfseries\small,
      anchor=west
    ] at ([xshift=5pt,yshift=-1.5pt]interior.north west) {#1};
  },
}

\definecolor{pureincrease}{RGB}{190,45,45}
\definecolor{puredecrease}{RGB}{35,90,180}

\newcommand{\ourdata}{\textsc{PULSE}}

\title{Do LLMs Speak the User’s Language?\\ A Study of Personalized Language Selection in Multilingual LLMs}

\author{
    Jaehoon Jang\textsuperscript{\textnormal{1}},~~Seunghee Lee\textsuperscript{\textnormal{1}},~~Sanghoon Yoo\textsuperscript{\textnormal{2}},~~Jisoo Mok\textsuperscript{\textnormal{2}}\thanks{Corresponding Author} \\ \\
\textsuperscript{\textnormal{1}}Hankuk University of Foreign Studies~~~~\textsuperscript{\textnormal{2}}Hanyang University \\ \\
\texttt{\{hooni,shamy0527\}@hufs.ac.kr,~\{shyoo,jisoomok\}@hanyang.ac.kr}}

\begin{document}
\maketitle

\begin{abstract}
Large Language Models increasingly interact with multilingual users, yet personalization of LLMs' response-languages remains overlooked. 
We present the first systematic study of user-aware response-language generation: customizing response-languages to users' linguistic background inferred from prior interactions.
To this end, we introduce~\textbf{\ourdata}, a benchmark of code-switched dialogue histories, queries, and user-conditioned reference responses, with linguistically grounded automated evaluation. 
Experiments reveal that LLMs are driven by surface cues in the query, often responding in a user-inappropriate form.
To mitigate this challenge, we propose a two-stage alignment pipeline, comprised of supervised grounding on dialogue-derived user cues and reinforcement-learning post-training with user-aware rewards.
Our dataset and code are released at~\url{https://github.com/LaMI-Labs/PULSE}
\end{abstract}

\section{Introduction}

Large Language Models (LLMs) are increasingly deployed as conversational assistants across a wide range of applications, making it important to understand how effectively they support human users~\cite{ou2024dialogbench, zhao2025personalens}. 
Despite the expanding scope of LLMs' applications, existing benchmarking of LLMs is largely focused on task accuracy in isolated question-answering settings~\cite{hendrycks2020measuring,cobbe2021training,srivastava2023beyond}. 
When LLMs engage in sustained interaction with humans, however, additional challenges that are not readily captured by such single-turn evaluations emerge. 
In particular, failure to retain user-relevant information or remain consistent with prior interactions can undermine users' engagement~\cite{cox2023comparing,maharana2024evaluating,tan2025prospect} and trust~\cite{baughan2023mixed,landerberg2026trust} in future interactions.
Consequently, personalization, the ability to adapt model behavior and responses to individual users, has become an increasingly important research direction for improving the quality of user experience and overall satisfaction~\cite{chen2024recent,westhausser2025enabling}.

Prior works on personalization revolved around customizing the \textit{content} of the model response to the users~\cite{cheng2024dialogues,liu2025llms+}. 
In contrast, the question of \textit{in what linguistic form} that content should be delivered in a user-tailored manner has received relatively limited attention~\cite{blevins2026language}.
For instance, a multilingual user may prefer general explanations in their native language while being more familiar with technical terminology in their secondary language, making a carefully code-switched response more natural and comprehensible than a fully monolingual one~\cite{brixey2026can}.
In this work, we study~\textit{user-aware response-language generation}, the task of selecting and combining languages in a response based on an individual user's linguistic background.
Successfully carrying out this task enables LLMs to support natural and socially appropriate interactions by conveying language-specific knowledge and cultural nuances more effectively.

User-aware response-language generation differs from conventional content personalization in two respects. 
First, the LLM must infer the user’s linguistic background from behavioral patterns in their prior interactions,~\textit{e.g.,} user's language choice,  or lexical usage.
This inference is non-trivial because such attributes are rarely stated explicitly and must instead be recovered from indirect signals~\cite{mok2025exploring}. 
Second, the LLMs must flexibly turn this understanding into the most appropriate response-language while preserving semantic fidelity, fluency, and discourse coherence.
This process is further complicated by the limitation of current LLMs in producing natural code-switched text~\cite{kuwanto2026linguistics}. 

\begin{figure*}[t]

  \includegraphics[width=\textwidth]{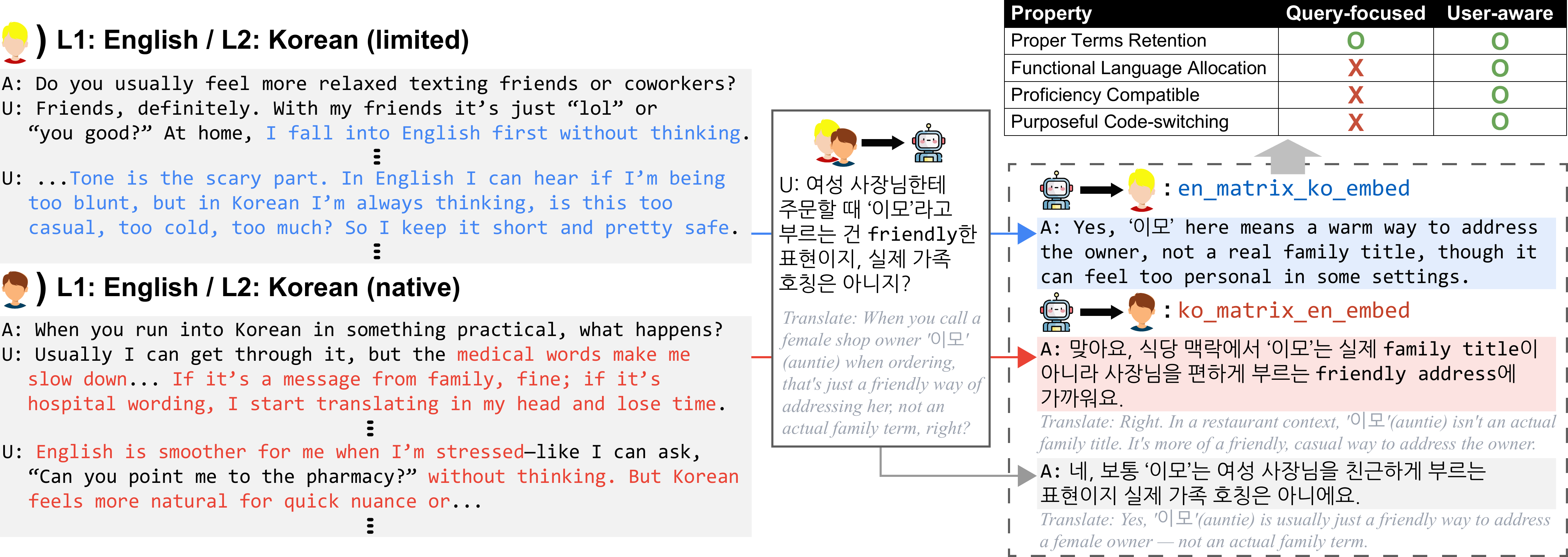}
  \caption {Illustration of user-aware response-language generation. Given the same Korean--English code-switched query, the query-focused response (Gray) follows the surface form of the query, whereas the user-aware response (Red and Blue) adapts the language of the explanation to the user's Korean proficiency inferred from prior dialogue while preserving the culture-specific term \textit{imo} (`auntie').}
  \label{fig1:example}
  \vspace{-1em}
\end{figure*}

To systematically investigate aforementioned challenges, we introduce~\textbf{\ourdata}~(\textbf{P}ersonalized \textbf{U}ser-aware \textbf{L}anguage \textbf{S}election \textbf{E}valuation), a new synthetic benchmark for user-aware response-language generation. 
Multi-turn, code-switched dialogue histories in~\ourdata~implicitly reveal a multilingual user’s linguistic background. 
Each dialogue history in~\ourdata~is paired with a code-switched query and a reference response whose linguistic form is conditioned on the user information.
By presenting the same query under different user profiles, this design enables us to isolate whether models adapt their response-language to information revealed through dialogue history rather than merely following the linguistic form of the query. 
In addition,~\ourdata~includes an automatic evaluation process that assesses the appropriateness of response-language forms using linguistically motivated criteria, eliminating the need for LLM-based judges or manual inspection.

Our initial exploration with~\ourdata~reveals that response-languages of current LLMs are contingent on the surface-level linguistic structure of the final query, failing to deliver the response in user-appropriate form.
Unfortunately, a code-switched query may conflict with the user's proficiency in each language or their preferred language for extended explanations. 
This finding highlights a key blind spot of previous query-centered approaches to response-language selection~\cite{oh2026ola}; as illustrated in~\figurename~\ref{fig1:example}, meaningful personalization requires LLMs to look beyond the most recent utterance and draw on interaction history to infer user-specific linguistic characteristics.

Motivated by this observation, we propose a holistic two-stage alignment pipeline to encourage personalization of response-language to users’ linguistic backgrounds. 
The first phase employs supervised fine-tuning (SFT) with reference responses to ground response generation in implicit linguistic cues embedded in the dialogue history. 
The subsequent reinforcement-learning(RL)-based post-training phase with graded, proficiency-aware rewards enables the model to learn a generalizable policy that favors response-language forms compatible with the user.
Extensive experimental results demonstrate that the proposed alignment protocol effectively induces proficiency-aware policy for user-aware response-language selection in LLMs.
\section{Related Work}

\subsection{Personalization of LLMs}
\label{relate2.1}
Personalization, the process of tailoring LLMs' responses to individual users, plays a crucial role in the user experience of LLM-based conversational systems~\citep{tseng2024two}. 
Although personalization has become increasingly important, even frontier LLMs often fail to tailor their responses appropriately to individual users~\citep{jiang2025know}.
Early works in personalization aimed to condition the generation process on persona statements or user-specific records that were directly available in the context~\citep{zhang2018personalizing, salemi2024lamp}.
Recent works have begun to explore more realistic and practical scenarios in which the model must infer them from its previous interactions with the user~\citep{zhao2025llms, zhong2025evaluating}.
Similarly, our work explores whether LLMs can adapt their responses to a user's linguistic background, which can be inferred from previous dialogue turns.

To the best of our knowledge, our work is the first to extend personalization beyond the~\emph{content} of a response to its~\emph{linguistic form}, adapting how LLMs' responses are styled to the user’s background. 
Most closely related to our work,~\citet{xu2026controlling} customize the linguistic difficulty of responses to the proficiency of second-language learners. 
Their approach, however, modulates complexity within a single, predetermined target language rather than selecting or combining languages according to the user’s linguistic background.
\subsection{Code-Switching for Response-Language Generation}
Code-switching (CS), the alternation between two or more languages within a discourse, is a natural form of communication among multilingual speakers~\citep{poplack1980sometimes}.
Early code-switching research focused mostly on natural language understanding, evaluating language models' ability to process code-switched input~\citep{khanuja2020gluecos,winata2021multilingual,prasad2021effectiveness}.

Recent studies have begun to examine whether LLMs can generate fluent and linguistically plausible code-switched text~\citep{yong2023prompting,kuwanto2026linguistics}.
Yet, these studies primarily evaluate models' general code-switching generation capabilities and the linguistic quality of their outputs, overlooking which linguistic form is appropriate in a particular interaction.
\citet{oh2026ola} evaluate whether LLMs can infer the implicitly expected output language from cues in the current query, but their query-centered formulation does not account for the possibility that the same query may warrant different response forms for different users.
Our work substantially extends query-centered output-language alignment to user-aware response generation by studying if LLMs can generate an appropriate monolingual or code-switched response according to each user's language proficiency.

\begin{table*}[t]
\centering
\renewcommand{\arraystretch}{0.95}
\resizebox{1\linewidth}{!}{
\begin{NiceTabular}{l|ccc|cccc|c}
\toprule
\multirow{2}{*}{Data Split}
& \multicolumn{3}{c|}{Number}
& \multicolumn{4}{c|}{Unique Query Type}
& \multirow{2}{*}{Instances} \\
\cmidrule{2-8}
& Personas & Dialogues & Queries
& Gen. (ko) & Gen. (en) & Cul. (ko/zh) & Cul. (en)
& \\
\midrule
Train
& 6 & 42 & 659
& 198 & 181 & 140 & 140
& 27,678 \\
Test (Seen User) 
& 6 & 42 & 282
& 85 & 77 & 60 & 60
& 11,844 \\
Test (Unseen User)
& 6 & 18 & 282
& 85 & 77 & 60 & 60
& 5,076 \\
Test (Unseen Language)
& 6 & 60 & 120
& - & - & 60 & 60
& 7,200 \\
\bottomrule
\end{NiceTabular}
}
\vspace{-0.3em}
\caption{Dataset configuration of~\ourdata. Korean--English data are divided into train, seen user test, and unseen user test sets, while Chinese--English data form a separate test set to study unseen-language transfer.
Gen. and Cul. denote general and cultural queries, respectively, and language codes in parentheses indicate the matrix language.}
\label{tab:pulse_koen_statistics}
\end{table*}

\section{\ourdata: Dataset Construction \& Configuration}
\label{sec3:pulse}


To investigate user-dependent variations in bilingual communication, we introduce~\ourdata, a benchmark that pairs dialogue histories containing implicit evidence of a user’s proficiency in primary and secondary languages (L1 and L2) with code-switched queries and controlled reference responses.
Sections~3.1 and~3.2 describe how the persona-conditioned dialogue histories and query--reference response sets in~\ourdata~are constructed.

\subsection{Persona-Conditioned Dialogues}

\paragraph{Bilingual Persona Design.}
Each bilingual user in~\ourdata~is characterized by their L1 and their proficiency in L2.
We consider three levels of L2 proficiency: Limited, Working, and Native. 
For each supported language pair, either language may serve as the user's L1, with the other serving as L2.
Combining the two possible L1--L2 settings with the three proficiency levels yields six bilingual persona configurations per language pair, as summarized in Table~\ref{tab:personas}.
The current version of~\ourdata~instantiates this persona design for Korean--English and Chinese--English bilingual users.
Using the same persona schema across both language pairs allows us to evaluate whether learned personalization behavior transfers to a previously unseen bilingual setting.

\paragraph{Synthetic Dialogue Generation.}
For each of the six bilingual persona configurations in a language pair, we generate ten multi-turn dialogue histories, yielding 60 distinct histories per language pair; each dialogue history thus represents a distinct realization of a synthetic user.
The prompt template used for dialogue generation is provided in Figures~\ref{fig:prompt_dialogue_generation_1}--\ref{fig:prompt_dialogue_generation_2}.
The prompt is designed to reveal the user's linguistic background indirectly through their behavior across multiple turns,~\textit{e.g.,} requests for shorter or simpler explanations or difficulty understanding idioms or humor.
The dialogue histories therefore provide a controlled setting in which LLMs must infer a user's linguistic background from interactional evidence and use that understanding when responding to a subsequent query.

\subsection{Code-Switched Query-Response Pairs}
\label{sec:query-response-pairs}
\ourdata~provides two types of code-switched queries: general and cultural queries.
General queries test whether a model can prioritize user-level evidence over the surface linguistic form of the current query.
Cultural queries introduce an additional content-level constraint, testing whether a model can preserve language-bound cultural expressions while still leveraging an appropriate language for the surrounding explanation.

\paragraph{General Code-Switched Queries.}
For general queries, we source two Korean--English code-switched variants from OLA~\cite{oh2026ola}: \texttt{ko\_matrix\_en\_embed} and \texttt{en\_matrix\_ko\_embed}.
The matrix language governs the query’s primary grammatical structure, while the embedded language supplies shorter expressions inserted into the matrix~\cite{myers1997duelling}. 
In OLA, the linguistic structure of a query is treated automatically as the expected response-language.
In~\ourdata, by contrast, the influence of the query form may be overridden when an alternative response-language is more accessible or appropriate for the user.

\paragraph{Culturally-Grounded Code-Switched Queries.}
In bilingual communication, culturally specific expressions or proper names may lose part of their connotation when translated.
At the same time, preserving such an expression in its original language does not necessarily imply that the entire response should be provided in that language.
The cultural queries of~\ourdata~therefore require LLMs to select appropriate response-languages at two different levels.
At the local level, a culturally specific expression may need to remain in its original language.
At the global level, the matrix language of the explanation should remain appropriate for the individual user's linguistic background and proficiency.
We construct Korean--English and Chinese--English cultural query sets following the process detailed in Section~\ref{app:cultural-query-construction}.
The Korean--English cultural queries complement the general query set by introducing content-sensitive language choices.
The Chinese--English cultural queries additionally support evaluation on a held-out language pair.

\paragraph{Reference Response Construction.}
\label{sec:reference_response_construction}
For each dialogue history and compatible query, we first determine the matrix language of a reference response according to the underlying persona.
Let \(p\) denote a user persona, \(q\) a code-switched query, and \(z(p)\) the user's L2 proficiency.
The target matrix $m^*$ of a reference response $y^*$ is defined to be:
\begin{equation}
\resizebox{0.98\columnwidth}{!}{$\displaystyle
m^{*}(y^*) =
\begin{cases}
L_{1}(p),
&
z(p)\in\{\textsc{Limited},\textsc{Working}\},
\\[1mm]
m(q),
&
z(p)=\textsc{Native},
\end{cases}
$}
\label{eq:target_matrix}
\end{equation}
where \(L_{1}(p)\) is the user's L1 and \(m(q)\) is the matrix language of the current query.

This assignment captures the assumption that users with Limited or Working L2 proficiency may understand short L2 expressions, while still finding extended or nuanced explanations easier to process in their L1.
On the other hand, for users with Native L2 proficiency, the target matrix of response-language follows that of the current query because these users are assumed to process both languages comfortably.
Once the target matrix language is determined, the other language is naturally used as the embedded language.
Afterwards, we generate a one-sentence reference response that answers the query in the assigned response-language through the process in Section~\ref{app:ref_response_gen}.
Note that reference responses are intended to provide one frame of reference for determining an appropriate response-language form for bilingual users, rather than defining the only acceptable form.

The final Korean--English dialogue--query--response triplets are partitioned into a train set and two test sets: Seen Dialogue (User) | Unseen Query and Unseen Dialogue (User) | Unseen Query settings.
The Chinese--English dialogue--cultural query--response triplets provide a test for cross-lingual transfer, requiring models to apply personalization behavior learned from Korean--English data to an unseen language pair.
The full dataset configuration is summarized in~\tablename~\ref{tab:pulse_koen_statistics}.

\begin{figure*}[t]
  \centering
  \includegraphics[
    width=\textwidth
]{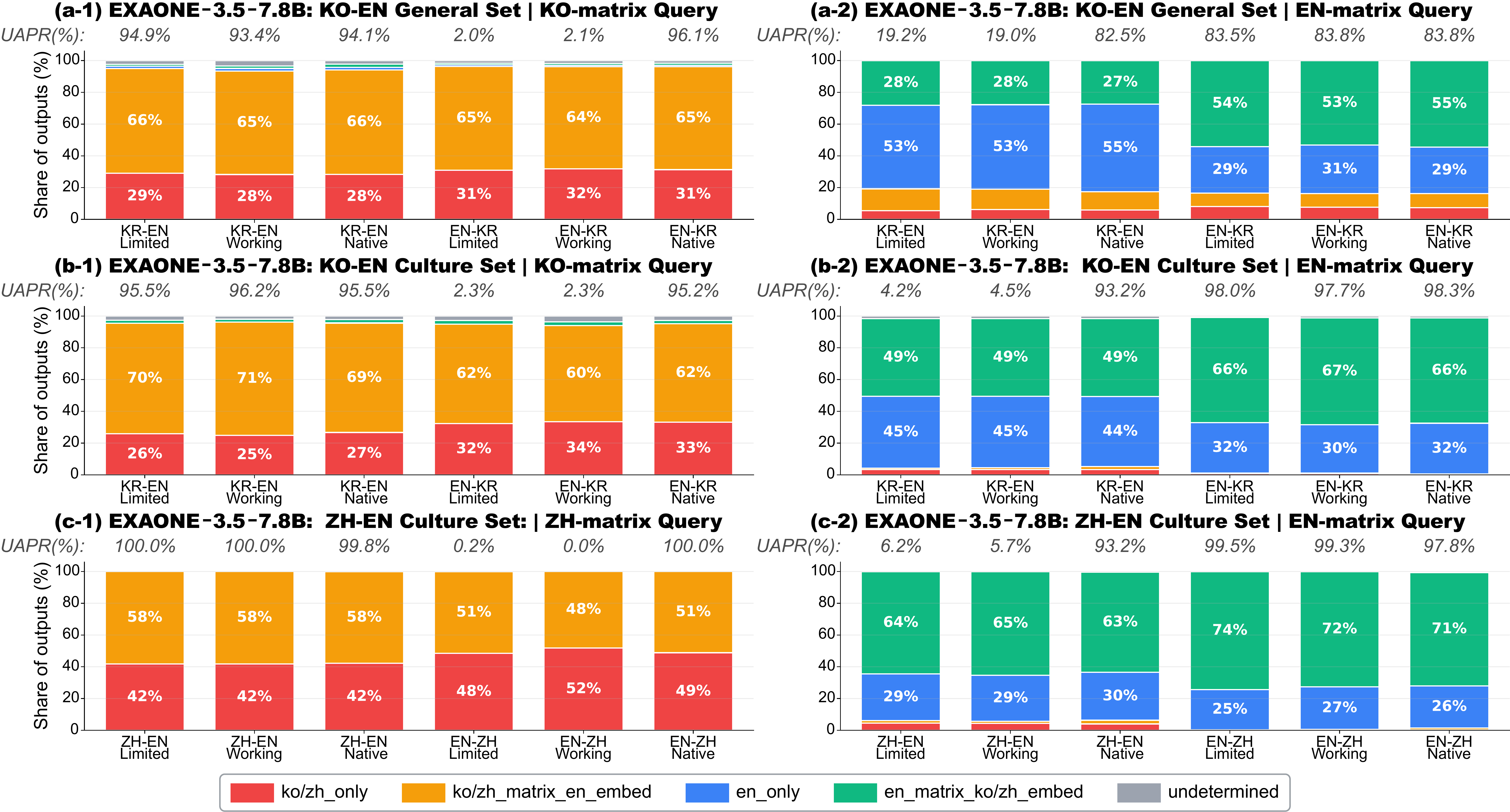}
  \caption{Response-language distributions and persona-level UAPR of the EXAONE-3.5-7.8B model. It is clearly visible that without additional alignment, EXAONE struggles to provide responses in user-aware response-language.}
  \label{fig2:motivation}
  \vspace{-5pt}
\end{figure*}

\section{Preliminary Analysis}

\subsection{Automated Evaluation Protocol}
\label{sec:evaluation_process}


\paragraph{Linguistically-Motivated Response-Language Detection.}
To enable scalable evaluation, we develop a linguistically motivated process for automatically identifying response-language forms.
The detection pipeline consists of three stages: language presence detection, grammatical signal extraction, and matrix language decision.
Rather than relying solely on token-level language proportions, our pipeline uses language-specific morphological and syntactic cues to determine which language provides the primary grammatical frame of a code-switched response, thereby serving as the matrix language.
Details of the proposed detection process are provided in Section~\ref{app:response_form_detector}.
For Korean--English, it assigns each response to one of: \texttt{ko\_only}, \texttt{en\_only}, \texttt{ko\_matrix\_en\_embed}, \texttt{en\_matrix\_ko\_embed}, or \texttt{undetermined}; analogous labels are used for Chinese--English.

To validate the detector, we asked two linguistics experts fluent in both languages to independently label 100 responses.
The two experts agreed on 95.0\% of responses (Cohen's $\kappa=0.933$).
Detector--expert agreement on the full sample was 75--77\% ($\kappa=0.688$--$0.713$), with most disagreements involving \texttt{undetermined} outputs.
For the 80 responses assigned a form by the detector, agreement with each expert was 93.75\% ($\kappa=0.917$).
These results support using the detector to identify response forms in our evaluation (see Appendix~\ref{app:human_study_detector} for further details).


\paragraph{User-Aware Pass Rate~(UAPR).} 
We then quantify whether the detected form of response-language is appropriate for the corresponding user using User-Aware Pass Rate (UAPR), computed as: $\mathrm{UAPR}
=
\frac{N_{\mathrm{pass}}}{N_{\mathrm{total}}}$, where \(N_{\mathrm{pass}}\) and \(N_{\mathrm{total}}\) denote the number of responses classified into user-appropriate forms and the total number of evaluated responses, respectively.

We operationalize user appropriateness based on the user's L1 and L2 proficiency. 
For users with Limited or Working L2 proficiency, a response is considered appropriate when its detected matrix language matches the user's L1. 
Both an L1-only response and an L1-matrix code-switched response are therefore accepted. 
For those with Native L2 proficiency, both L1 and L2 are treated as appropriate in our evaluation, and accepted response-forms are determined by the matrix language of the current query.
Responses classified as \texttt{undetermined} are treated as failures.
Evaluation criteria for Korean--English and Chinese--English pairs are summarized in Tables~\ref{tab:uapr_pass_criteria_ko} and \ref{tab:uapr_pass_criteria_zh}.

To assess the UAPR pass criterion, we asked 25 Korean--English bilingual speakers to select suitable responses for each user profile and query.
Of their 1,948 selections, 1,375 (70.6\%) met the criterion.
This proportion was highest for profiles with Limited L2 proficiency (81.4\%).
Table~\ref{tab:pulse-cr-policy} reports the results by L2 proficiency (see Appendix~\ref{app:human_study_survey} for further details).



\subsection{Response-Language Misalignment in Multilingual LLMs}
\label{subsec:howselect}
Using~\ourdata~and the automated evaluation protocol, we analyze response-language selection behavior of EXAONE-3.5-7.8B-Instruct~\cite{exaone-3.5} and Qwen-3.5-9B~\cite{qwen3.5}, two popularly-used multilingual open-source LLMs.
Figure~\ref{fig2:motivation} presents representative results from EXAONE; the results on Qwen are reported in~\figurename~\ref{fig:appendix_qwen_base} due to the page constraint. 
Details of the generation setting are provided in~Section~\ref{app:inference_settings}, with corresponding prompt templates in~\figurename~\ref{fig:prompt_inference_evaluation}.

\begin{keyfinding}{Key Finding 1}
LLMs select their response-language primarily based on the current query’s matrix language and largely overlook user-specific linguistic cues, resulting in poor personalization of response-language.
\vspace{-0.4em}
\end{keyfinding}

They both show that LLMs prioritize the linguistic structure of the most recent query over the user’s linguistic background when selecting their response-language form. 
For Korean-matrix general queries, EXAONE predominantly generates~\texttt{ko\_matrix\_en\_embed} responses, resulting in UAPR scores of only 2.0\% and 2.1\% for EN--KR Limited and Working users, respectively. 
A similar pattern emerges for English-matrix queries: because EXAONE's response-language largely defaults to \texttt{en\_matrix\_ko\_embed}, its UAPR scores reach only 19.2\% and 19.0\% for KR--EN Limited and Working users, respectively. 
This query-dependent behavior persists across in Qwen as well, with the response-form distribution closely following the matrix language of the current query.

\begin{keyfinding}{Key Finding 2}
LLMs show weak sensitivity to users’ dominant language direction and fail to adapt their response-language to fine-grained differences in proficiency.
\vspace{-0.4em}
\end{keyfinding}
We observe a modest increase in responses that follow the query’s matrix when it matches the user’s L1. 
For instance, on English-matrix queries, EXAONE produces more~\texttt{en\_matrix\_ko\_embed} or \texttt{en\_matrix\_zh\_embed} responses for English-L1 users than for Korean/Chinese-L1 users. 
This suggests that EXAONE can extract coarse cues about the user’s proficiency from the dialogue history, but these cues have only a marginal effect on its response-languages.

More importantly, LLMs do not distinguish between users with different degrees of L2 proficiency.
Within the same language direction and query condition, EXAONE and Qwen exhibit nearly identical response-language distributions on Limited, Working, and Native users.
This pattern holds for both general and cultural queries and is also observed in the Chinese--English test set, indicating that it is neither query-specific nor confined to Korean--English interactions.
Collectively, our preliminary experiments reveal a clear gap between current LLMs’ response-language behavior and the fine-grained adaptation required for user-aware multilingual interaction, motivating the alignment approach introduced in the following section.

\newcommand{\allnum}[1]{%
  \multicolumn{1}{@{\hspace{1.5em}}r}{#1}%
}
\begin{table*}[t]
\centering
\renewcommand{\arraystretch}{0.95}
\resizebox{\linewidth}{!}{
\begin{NiceTabular}{l|l|rrrrr|rrrrr}
\toprule
\multirow{2}{*}{\textbf{Model}}
& \multirow{2}{*}{\textbf{Method}}
& \multicolumn{5}{c|}{\textbf{UAPR(\%): Seen Dialogue}}
& \multicolumn{5}{c}{\textbf{UAPR(\%): Unseen Dialogue}} \\
\cmidrule{3-12}
&
& \multicolumn{1}{c}{\textbf{Gen.(ko)}}
& \multicolumn{1}{c}{\textbf{Gen.(en)}}
& \multicolumn{1}{c}{\textbf{Cul.(ko)}}
& \multicolumn{1}{c}{\textbf{Cul.(en)}}
& \multicolumn{1}{c|}{\textbf{All}}
& \multicolumn{1}{c}{\textbf{Gen.(ko)}}
& \multicolumn{1}{c}{\textbf{Gen.(en)}}
& \multicolumn{1}{c}{\textbf{Cul.(ko)}}
& \multicolumn{1}{c}{\textbf{Cul.(en)}}
& \multicolumn{1}{c}{\textbf{All}} \\
\bottomrule
\toprule

\Block{4-1}{EXAONE}
&
Base
& 63.6
& 62.2
& 64.5
& 66.2
& \allnum{64.0}
& 64.2
& 61.3
& 64.4
& 65.5
& \allnum{63.7}
\\

&
SFT
& 89.0
& 92.0
& 97.9
& 99.1
& \allnum{93.8}
& 78.4
& 83.4
& \textbf{83.3}
& 87.6
& \allnum{82.8}
\\

&
GRPO
& 61.9
& 63.8
& 62.9
& 66.7
& \allnum{63.7}
& 61.8
& 63.1
& 63.8
& 65.8
& \allnum{63.5}
\\

&
Ours.
& \textbf{97.9}
& \textbf{96.9}
& \textbf{99.4}
& \textbf{100.0}
& \allnum{\textbf{98.4}}
& \textbf{83.2}
& \textbf{87.6}
& 82.9
& \textbf{88.1}
& \allnum{\textbf{85.4}}
\\

\midrule

\Block{4-1}{Qwen}
&
Base
& 64.8
& 64.8
& 63.4
& 67.3
& \allnum{65.0}
& 65.0
& 64.0
& 63.2
& 66.9
& \allnum{64.8}
\\

&
SFT
& 95.3
& 94.7
& \textbf{100.0}
& \textbf{98.9}
& \allnum{96.9}
& 81.7
& 83.5
& 82.2
& 89.4
& \allnum{83.9}
\\

&
GRPO
& 62.5
& 74.7
& 59.0
& 77.6
& \allnum{68.3}
& 62.8
& 72.6
& 58.9
& 75.6
& \allnum{67.4}
\\

&
Ours.
& \textbf{99.3}
& \textbf{98.8}
& 99.4
& 98.6
& \allnum{\textbf{99.0}}
& \textbf{86.1}
& \textbf{86.6}
& \textbf{84.0}
& \textbf{89.7}
& \allnum{\textbf{86.6}}
\\

\bottomrule
\end{NiceTabular}%
}

\caption{UAPR comparison on Korean--English test sets. Gen. and Cul. denote general and cultural queries, and language codes in parentheses indicate the matrix language. The best result in each column is shown in bold.}
\label{tab:main_uapr}
\end{table*}

\begin{figure*}[t]
    \centering
    \includegraphics[width=\textwidth]
    {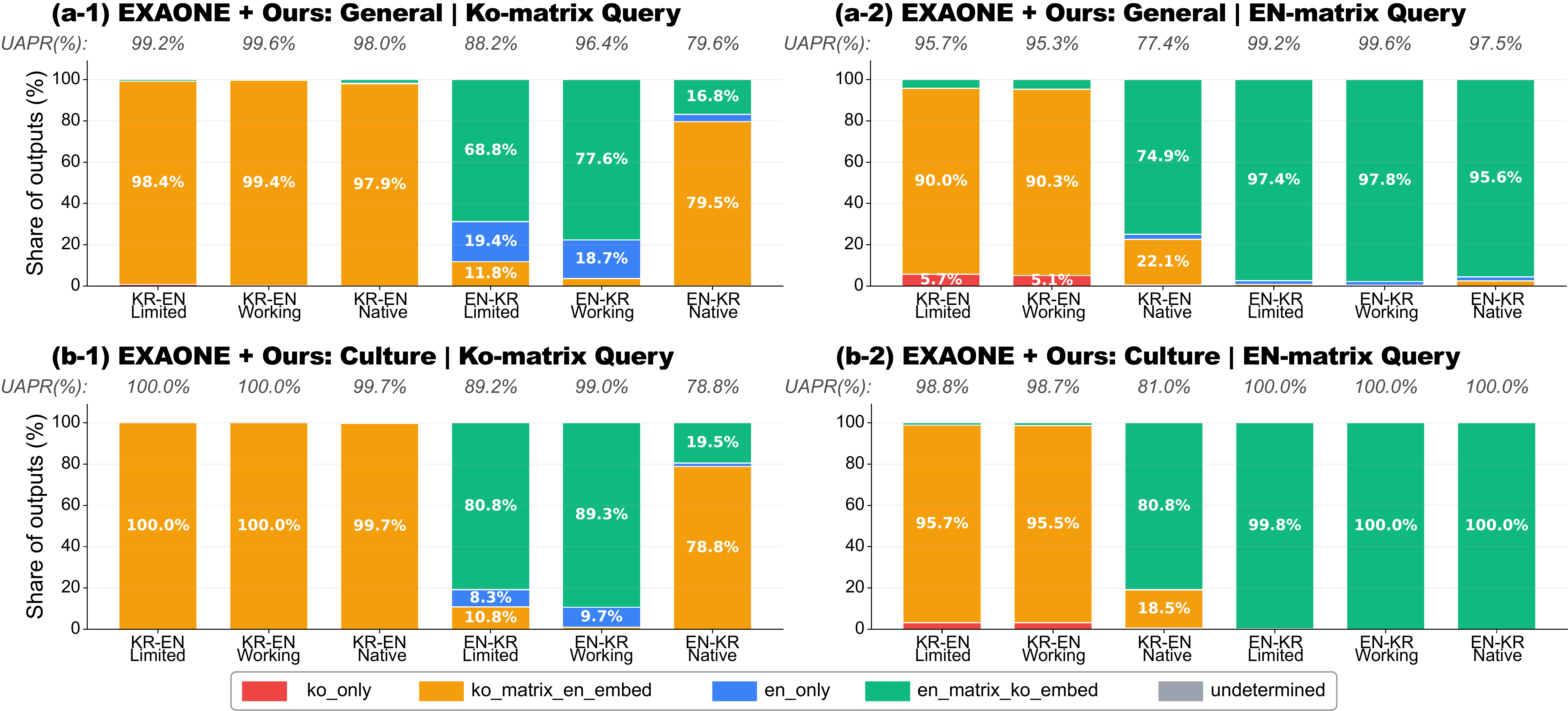}
    \caption{
    Response-language distributions of EXAONE aligned with our proposed method on the two Korean--English test sets combined.
    Persona-level UAPR is reported above each bar.}
    \vspace{-1em}
    \label{fig:response_form_distribution}
\end{figure*}

\section{User-Aware Response-Language Alignment}
\label{sec:alignment}


\subsection{Response-Language Grounding}
\label{sec:sft_alignment}
We propose a two-stage alignment pipeline that improves LLMs' ability to personalize their response-language form to the user’s linguistic background. 
The first stage employs supervised fine-tuning (SFT) to ground response generation in the implicit linguistic cues conveyed through the dialogue history. 
Given the Korean--English dialogue history and current query as input, we fine-tune the LLM to generate the corresponding reference response. 
Full training details are provided in Section~\ref{app:sft_settings}.

\subsection{UAPR-Guided Post-Training}
\label{sec:grpo_alignment}

Following SFT-driven grounding, we introduce RL-based post-training with a customized reward function to learn a generalizable policy for selecting user-appropriate response-language forms among multiple plausible outputs. 
Our reward function is designed to assign graded preferences to candidate responses according to their compatibility with the user and query conditions.
Let \(\hat{f}(y)\) denote the response-language detected for a candidate response \(y\), following the process described in Section~\ref{sec:evaluation_process}.
For brevity, let \(m^{*}=m^{*}(y^*)\), \(z=z(p)\), and let \(\bar{m}^{*}\) denote the non-target matrix,~\textit{i.e.,} the language other than \(m^{*}\).
Our fine-grained UAPR-driven reward is defined as:

\begin{equation}
\resizebox{0.98\columnwidth}{!}{$\displaystyle
R_{\mathrm{UAPR}}(y;p,q)
=
\begin{cases}
1.0,
& \hat{f}(y)=\mathrm{CS}(m^{*}),
\\[1mm]
r_{\mathrm{mono}}^{\mathrm{tgt}}(z),
& \hat{f}(y)=\mathrm{Mono}(m^{*}),
\\[1mm]
r_{\mathrm{CS}}^{\mathrm{ntgt}}(z),
& \hat{f}(y)=\mathrm{CS}(\bar{m}^{*}),
\\[1mm]
r_{\mathrm{mono}}^{\mathrm{ntgt}}(z),
& \hat{f}(y)=\mathrm{Mono}(\bar{m}^{*}),
\\[1mm]
0,
& \hat{f}(y)=\texttt{undetermined}.
\end{cases}
$}
\label{eq:uapr_reward}
\end{equation}

Here, \(\mathrm{CS}(m)\) denotes a code-switched response-language whose matrix language is \(m\), while \(\mathrm{Mono}(m)\) denotes a monolingual response in \(m\).
The superscripts \(\mathrm{tgt}\) and \(\mathrm{ntgt}\) distinguish response-forms that follow the target and non-target language directions, respectively.
Accordingly, \(r_{\mathrm{mono}}^{\mathrm{tgt}}(z)\), \(r_{\mathrm{CS}}^{\mathrm{ntgt}}(z)\), and \(r_{\mathrm{mono}}^{\mathrm{ntgt}}(z)\) specify the proficiency-conditioned rewards for a target-language monolingual response, a non-target-matrix code-switched response, and a non-target-language monolingual response, respectively.

In our reward design, code-switched responses with the target matrix of a reference response receive the maximum reward because it most directly realizes the response form appropriate for the given user and query.
Alternative forms receive partial rewards according to two factors: whether they preserve the target language direction and how compatible they are with the user’s L2 proficiency.
The reward assignment procedure is further detailed in Section~\ref{app:reward_assignments}.
We optimize LLMs under this reward using Group Relative Policy Optimization (GRPO)~\cite{shao2024deepseekmath}, whose implementation details are provided in~Section~\ref{app:grpo_hyperparam}.

\section{Results}
\label{sec:results}

We verify the effectiveness of the proposed alignment method on EXAONE-3.5-7.8B-Instruct~\cite{exaone-3.5} and Qwen-3.5-9B~\cite{qwen3.5}. 
We compare four approaches: Base, which uses open-source models as released without any additional training; supervised fine-tuning (SFT); GRPO initialized from the Base model (GRPO); and the Base model aligned with our full user-aware alignment pipeline (Ours).
Implementation details and hyperparameters for all compared methods are provided in Section~\ref{app:alignment_experimental_details}.
Our implementation uses Hugging Face Transformers \cite{wolf2019huggingface} and experiments are conducted on NVIDIA H200, H100, and L40 GPUs.

\subsection{Effects of User-Aware Alignment}
\label{sec:overall_uapr}

Table~\ref{tab:main_uapr} reports UAPR on Korean--English test sets.
UAPR-guided post-training without supervised grounding yields inconsistent improvements.
The full alignment pipeline (Ours) achieves the best overall UAPR on both splits. 
For EXAONE, UAPR is 98.4\% on seen users and 85.4\% on unseen users.
For Qwen, it is 99.0\% and 86.6\%, respectively.
These results suggest that SFT-based grounding and UAPR-guided post-training play complementary roles.
Figure~\ref{fig:response_form_distribution}, shows EXAONE's response-form distributions after alignment.
For Limited and Working profiles, EXAONE predominantly selects a code-switched form whose matrix language matches the user's L1, even when the query uses the other matrix language.
For Native profiles, it tends to preserve the query's matrix language.
These shifts toward the preferred code-switched forms follow the proficiency-conditioned target policy in Section~\ref{sec:reference_response_construction}.
Qwen shows similar profile-dependent patterns (\figurename~\ref{fig:appendix_qwen_post_train}).

Table~\ref{tab:pulse-cr-language-use} reports human ratings of 24 aligned responses (see Appendix~\ref{app:human_study_survey} for details).
The four criteria were term retention, functional language allocation, proficiency compatibility, and purposeful code-switching.
The mean rating of 4.15 out of 5 suggests that user-aware alignment can also yield responses rated favorably for their language use.

\begin{figure}[t]
    \centering
    \includegraphics[width=\columnwidth]
    {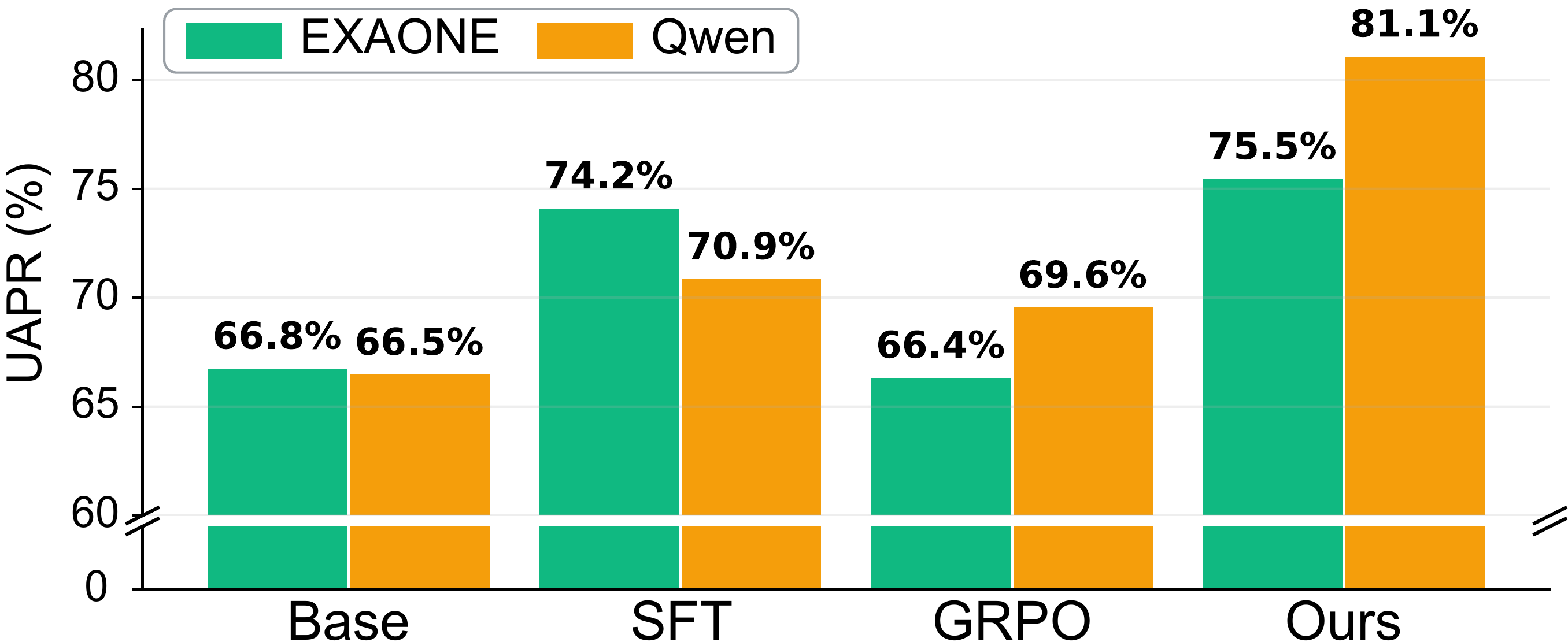}
    \caption{
    Cross-lingual transfer performance of EXAONE and QWEN aligned only with Korean--English data to Chinese--English cultural queries.}
    \label{fig:zh_transfer}
\end{figure}

\subsection{Cross-Lingual Transfer}
\label{sec:zh_transfer}

Figure~\ref{fig:zh_transfer} examines the cross-lingual transfer of response-language alignment learned from Korean--English interactions to Chinese--English users without additional language-pair-specific adaptation. 
After being trained with the proposed two-stage alignment protocol, EXAONE and Qwen both attain UAPRs of 75.5\% and 81.1\% on Chinese--English test set, indicating that the learned response-language personalization generalizes to unseen language pairs.


We also evaluate transfer to Korean--Chinese, a language pair without English.
We evaluate the models aligned on Korean--English without further training.
Relative to Base, UAPR increases from 55.1\% to 73.8\% for EXAONE and from 48.8\% to 72.9\% for Qwen (Table~\ref{tab:pulse-cr-transfer}).

\subsection{GRPO Reward and Training Data}
\label{sec:reward_analysis}

\begin{figure}[t]
    \centering
    \includegraphics[width=\columnwidth]
    {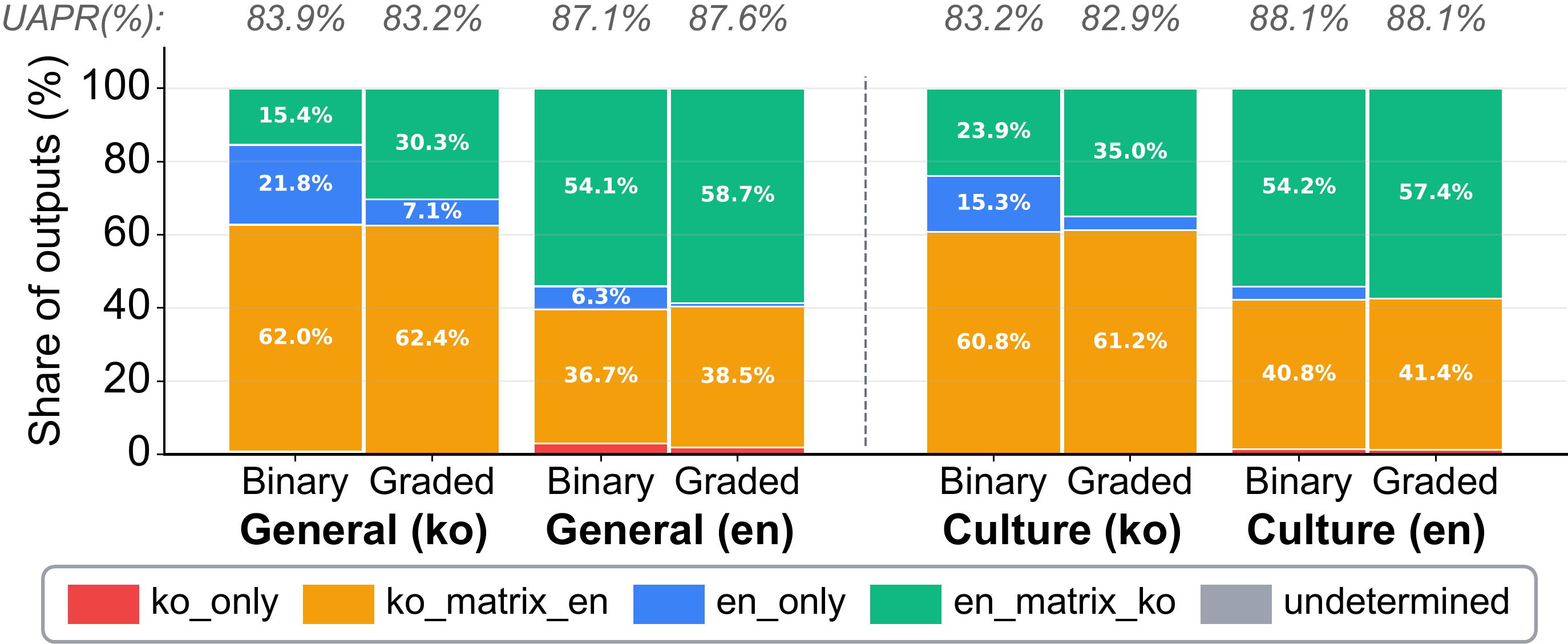}
    \caption{
    Effect of reward granularity on the alignment process. ``Binary'' and ``Graded'' denote the binary and fine-grained rewards, respectively.}
    \vspace{-1em}
    \label{fig:reward_granularity}
\end{figure}

\paragraph{Reward Granularity.} We compare our fine-grained, proficiency-conditioned reward with a binary UAPR reward that assigns 1 to passing response forms and 0 otherwise. 
\figurename~\ref{fig:reward_granularity} summarizes the comparison for EXAONE on unseen dialogues, with extended results in \figurename~\ref{fig:appendix_binary_reward}.
The two rewards differ by at most 0.7 percentage points in UAPR on each of the four query subsets.
At comparable UAPR, the fine-grained reward yields fewer monolingual and more code-switched responses, suggesting that it helps GRPO express finer preferences among acceptable forms.

\paragraph{Proportion of Native Training Examples.}
We next test whether increasing the share of Native examples helps EXAONE follow the query's matrix language when it differs from the user's L1.
We keep the graded reward and training set size fixed.
On unseen dialogues and queries, increasing the Native share from one-third to two-thirds raises UAPR for these Native cases from 38.8\% to 46.8\%.
UAPR for Limited and Working profiles combined falls from 92.7\% to 91.0\%.
Overall UAPR stays near 85\% (Table~\ref{tab:pulse-cr-training-mix} in Appendix~\ref{app:grpo_training_mix}).

\subsection{Prompting Baselines}
\label{sec:instruction_effect}

\begin{figure}[t]
    \centering
    \includegraphics[width=\columnwidth]
    {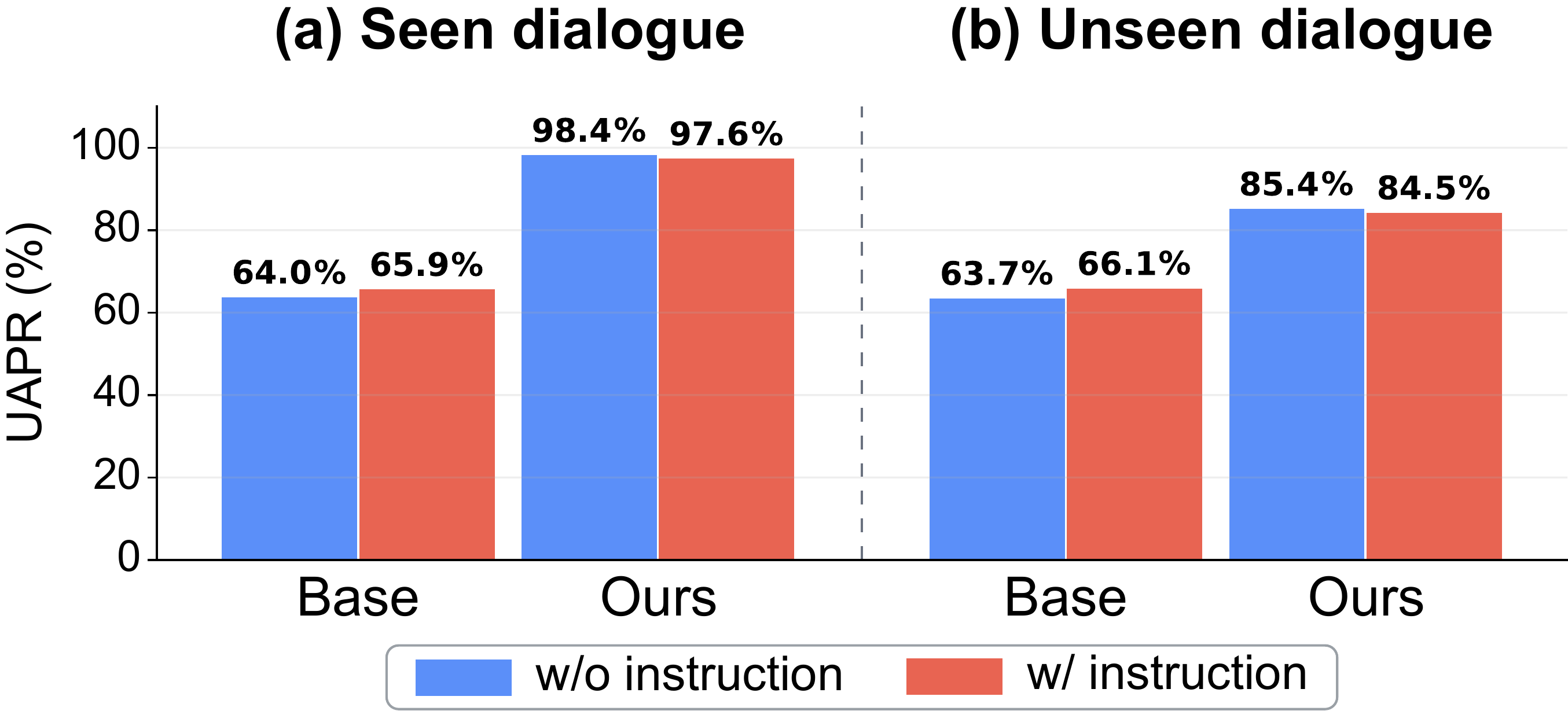}
    \caption{
    How the explicit task instruction affects the response-language of EXAONE.}
    \vspace{-1.5em}
    \label{fig:task_instruction}
\end{figure}

\paragraph{Task Instruction.}
\figurename~\ref{fig:task_instruction} compares EXAONE with and without the brief task instruction before and after alignment.
With the instruction, Base UAPR is 1.9--2.4 \%p higher across the two splits, and SFT+GRPO UAPR is 0.8--0.9 \%p lower.
\figurename~\ref{fig:appendix_qwen_fig4} shows the corresponding Qwen results.
Figures~\ref{fig:prompt_inference_evaluation} and~\ref{fig:prompt_sft} show the prompts used for inference and supervised fine-tuning, respectively.

\paragraph{Few-Shot and Profile-Based Prompting.}
Table~\ref{tab:pulse-cr-prompting} compares UAPR for the prompting
baselines and SFT+GRPO.
On unseen dialogues and queries, few-shot prompting~\citep{brown2020language} yields 61.1\% UAPR for EXAONE and 64.1\% for Qwen. 
With an explicit language selection policy, the two-stage baseline first infers the profile from dialogue history and
achieves 65.1\% and 71.9\%, respectively. 
With benchmark-assigned profiles, the scores are 68.3\% and 77.4\%.

GPT-5.6 Terra~\citep{openai2026gpt56} achieves 63.2\% when it infers the profile and applies the policy in one call. 
With the assigned profile, it reaches 78.2\%.
SFT+GRPO exceeds these prompting baselines, reaching 85.4\% for EXAONE and 86.6\% for Qwen.


\section{Conclusion}
In this work, we introduced~\ourdata, a systematic benchmark for investigating user-aware response-language generation from code-switched dialogue histories.
Using~\ourdata, we demonstrated that widely used LLMs remain limited in their ability to adapt response-language choices to users’ linguistic backgrounds, often relying on surface-level cues in the current query rather than users’ broader linguistic profiles.
To address this limitation, we further proposed a two-stage alignment pipeline that combines supervised grounding with proficiency-aware reinforcement learning. 
We hope our research provides a foundation for more adaptive and human-aligned multilingual assistants, capable of personalizing not only \textit{what} they communicate, but also \textit{how} they communicate with individual user.

\section*{Acknowledgments}
This work was supported by Institute of Information \& communications Technology Planning \& Evaluation (IITP) under the artificial intelligence semiconductor support program (IITP-2026-RS-2023-00253914) funded by the Korea government (MSIT); the Institute of Information \& communications Technology Planning \& Evaluation (IITP) grant funded by the Korea government (MSIT) (No.RS-2020-II201373, Artificial Intelligence Graduate School Program (Hanyang University)); Artificial intelligence industrial convergence cluster development project funded by the Ministry of Science and ICT(MSIT, Korea) \& Gwangju Metropolitan City; and the National Research Foundation (NRF) of Korea (No.RS-2026-25478246).

\section*{Limitations}

Human evaluation was limited to the Korean--English pair and involved bilinguals with Korean as their L1 and English as their L2.
PULSE's synthetic dialogue histories may underrepresent real-world bilingual variation, and its three L2 proficiency levels simplify a continuous, multidimensional construct and leave within-profile differences in response-language preferences unmodeled.
Native speakers or cultural experts should assess the cultural queries' contextual appropriateness.

Drawing on the Matrix Language Frame model's distinction between matrix and embedded languages, our detector focuses on intra-sentential word or phrase insertions.
Assigning a single matrix language per response can obscure switches between clauses or sentences.
The current implementation is limited to Korean, English, and Chinese.
Future work could extend the detector to capture a wider range of code-switching patterns and support additional languages.

\section*{Ethical Considerations}

Our synthetic dialogue histories contain no personally identifiable information from real users.
Inferring users' linguistic backgrounds and proficiency from real conversations may raise privacy and misclassification concerns.
Model generated cultural data may contain factual errors, oversimplifications, or stereotypical associations.






\bibliography{custom}

\clearpage
\appendix

\setcounter{table}{0}
\renewcommand{\thetable}{A\arabic{table}}
\setcounter{figure}{0}
\renewcommand{\thefigure}{A\arabic{figure}}
\setcounter{equation}{0}
\renewcommand{\theequation}{A\arabic{equation}}

\section{Dataset Construction Details}
\label{app:dataset_details}

\subsection{Bilingual Persona Configurations}
Table~\ref{tab:personas} summarizes the generic persona schema instantiated for the Korean--English and Chinese--English language pairs.

\begin{table}[h]
\centering
\small
\setlength{\tabcolsep}{4pt}
\renewcommand{\arraystretch}{1.05}

\begin{NiceTabular}{llll}
\toprule
\textbf{Persona Schema}
& \textbf{L1}
& \textbf{L2}
& \textbf{L2 Proficiency} \\
\bottomrule
\toprule
X\_EN\_LIMITED
& \(X\)
& English
& Limited \\
X\_EN\_WORKING
& \(X\)
& English
& Working \\
X\_EN\_NATIVE
& \(X\)
& English
& Native \\
EN\_X\_LIMITED
& English
& \(X\)
& Limited \\
EN\_X\_WORKING
& English
& \(X\)
& Working \\
EN\_X\_NATIVE
& English
& \(X\)
& Native \\
\bottomrule
\end{NiceTabular}

\caption{
Generic bilingual persona configurations instantiated for the Korean--English and Chinese--English language pairs.
Here, \(X\) denotes either Korean or Chinese.
Native denotes native or bilingual proficiency.
}
\label{tab:personas}
\end{table}

\subsection{Synthetic Dialogue Generation}
For each persona configuration, we generate ten dialogue histories
using GPT-5.4-mini\cite{openai2026gpt54mini}.
Each dialogue is generated based on a persona description, control metadata, and a scenario cue.
It consists of nine alternating user and assistant utterances,
beginning and ending with the user.
The prompt requires the user's linguistic background and approximate L2 proficiency to be revealed indirectly through concrete language-use habits, preferences, difficulties, and interactional experiences.
The complete prompt is presented in Figures~\ref{fig:prompt_dialogue_generation_1} and~\ref{fig:prompt_dialogue_generation_2}.

\subsection{Cultural Query Construction}
\label{app:cultural-query-construction}

This appendix describes the construction of the Korean--English and Chinese--English cultural query sets introduced in Section~\ref{sec3:pulse}.

The source-filtering and query-generation prompts are provided at the end of the section.

\subsubsection{Dataset Composition}
\label{app:cultural-query-composition}

We construct cultural query sets for the Korean--English and Chinese--English language pairs.
Each set contains two cultural components: a Korean or Chinese component and an Anglophone component. 
Each component contains 100 source items and two queries per source, one for each matrix-language direction.

The same 100 Anglophone source items are used in both language pairs, but their Korean--English and Chinese--English queries were generated separately. The resulting collection contains 800 queries derived from 300 unique source items.

\begin{table}[t]
\centering
\small
\setlength{\tabcolsep}{3.8pt}
\renewcommand{\arraystretch}{1.12}

\begin{NiceTabular}{llrr}[cell-space-limits=2pt]
\toprule
\textbf{Language Pair}
& \textbf{Source Culture}
& \textbf{Sources}
& \textbf{Queries} \\
\midrule
Korean--English
& Korean
& 100
& 200 \\
Korean--English
& Anglophone
& 100
& 200 \\
Chinese--English
& Chinese
& 100
& 200 \\
Chinese--English
& Anglophone
& 100
& 200 \\
\bottomrule
\end{NiceTabular}

\caption{
Composition of the cultural query sets.
The two language pairs share the same 100 Anglophone source items, but their code-switched queries were generated separately.
}
\label{tab:cultural-query-composition}
\end{table}

\subsubsection{Cultural Source Selection}
\label{app:cultural-source-selection}

Anglophone and Chinese cultural sources were drawn from MANGO~\citep{nguyen2024cultural}. 
The Anglophone pool covers the United States, the United Kingdom, and Canada, and the Chinese pool contains China-related items. 
Korean sources were drawn from K-Culture-Desc\citep{k_culture_2025}.

For the MANGO-based components, candidates are identified using culture labels, concept keywords, and regional metadata.
Each candidate received a 1--5 cultural-specificity score, a 1--5 code-switching query-potential score, and a \texttt{KEEP}, \texttt{BORDERLINE}, or \texttt{DROP} decision.
We retained 100 of 231 evaluated Anglophone candidates.
For the Chinese component, 102 of the 145 evaluated candidates were labeled \texttt{KEEP}, and 100 were selected for the final set.

K-Culture-Desc was filtered in two stages.
Of the original 530 items, 407 belonged to the nine retained categories.
The initial screening produced 363 \texttt{KEEP}, 36 \texttt{BORDERLINE}, and 8 \texttt{DROP} items.
The stricter second-stage filter produced 317 \texttt{KEEP}, 62 \texttt{BORDERLINE}, and 28 \texttt{DROP} items.
We formed a 200-source generation pool based on the filtering scores and category coverage. The later query-selection stage retained 100 of these sources.

\begin{table}[t]
\centering
\scriptsize
\setlength{\tabcolsep}{2.7pt}
\renewcommand{\arraystretch}{1.15}

\resizebox{\columnwidth}{!}{
\begin{NiceTabular}{lrrrrr}[cell-space-limits=2pt]
\toprule
\textbf{Source Pool}
& \textbf{Initial}
& \textbf{Pre-filtered}
& \textbf{Judged}
& \textbf{KEEP}
& \textbf{Final} \\
\midrule
Anglophone
& 16,700
& 841
& 231
& 100
& 100 \\
Chinese
& 4,332
& 381
& 145
& 102
& 100 \\
Korean
& 530
& 407
& 407
& 317
& 100 \\
\bottomrule
\end{NiceTabular}
}

\caption{
Cultural source selection statistics. Pre-filtering procedures differ across source datasets. For the Korean component, \textbf{Judged} and \textbf{KEEP} refer to the second-stage filtering results.
}
\label{tab:cultural-source-selection}
\end{table}

\subsubsection{Code-Switched Query Generation}
\label{app:cultural-query-generation}

For each non-English language \(X\), we used two query directions,
\texttt{X\_matrix\_en\_embed} and \texttt{en\_matrix\_X\_embed}, following the matrix-language notation defined in Section~\ref{sec:query-response-pairs}. The label indicates the intended direction only and does not require word-level substitution.

The Chinese cultural component generates one query directly for each source and direction.
The same procedure is used for the Chinese--English queries grounded in the shared Anglophone sources.

In contrast, the Korean cultural component generates three candidates per direction for each item in the 200-source generation pool.
A subsequent GPT-5.5\cite{openai2026gpt55} based selection stage retains 100 sources and selects one candidate per direction for each retained source.

For the Korean--English Anglophone component, three candidates per direction are generated for 87 sources, after which one candidate per direction is selected through manual review.
The remaining 13 sources are generated directly with one query per direction, and the two portions are then merged.

All generation prompts require the queries to remain faithful to the supplied cultural content and follow the specified matrix-language direction.
They also prohibit unsupported additions, literal question conversion, test-style wording, and repeated use of the same question frame.

\subsubsection{Quality Control}
\label{app:cultural-query-quality}

Generated records were programmatically checked for the expected matrix-language direction labels, required fields, and output counts.
The generation scripts also support retries after formatting failures and checkpoint-based resumption.

An initial pilot for the Chinese--English queries grounded in Anglophone sources showed that some Chinese-matrix queries copied long spans from the English source assertions.
We therefore revised the prompt to limit English insertions generally to one to three key expressions and to express country or location framing in the matrix language when natural.
We conducted a second pilot before full generation.

We also performed limited surface-level review.
For the Korean cultural component, five sources were replaced after a diversity review, and query selection was rerun for the replacement sources.
For the Chinese cultural component, three CJK--Latin spacing errors were corrected.
These checks do not constitute native-speaker or cultural-expert validation.

Cultural source filtering, query generation, and candidate selection primarily used GPT-5.5~\citep{openai2026gpt55} through the OpenAI Responses API with structured outputs.
The initial Korean source-screening stage used GPT-4o-mini~\citep{openai2024gpt4omini}, whereas the stricter filtering, query-generation, and query-selection stages used GPT-5.5.

\subsubsection{Generation Prompts}
\label{app:cultural-query-prompts}
Figures~\ref{fig:prompt_korean_filter_1}--\ref{fig:prompt_korean_filter_2} and Figure~\ref{fig:prompt_korean_query_generation} present the prompts used for cultural source filtering and code-switched query generation, respectively.
Figures~\ref{fig:prompt_chinese_filter_1}--\ref{fig:prompt_chinese_filter_2} and Figures~\ref{fig:prompt_chinese_query_generation_1}--%
\ref{fig:prompt_chinese_query_generation_3} present the corresponding prompts used for Chinese cultural source filtering and code-switched query generation.
We use language-specific versions of the generation prompt for Korean--English and Chinese--English, while maintaining the same high-level requirements for cultural-content preservation, matrix-language direction, and natural question formulation.


\subsection{Reference Response Generation}
\label{app:ref_response_gen}

For each dialogue--query pair, we assign a target response form based on the corresponding persona and query condition.
Using GPT-5.4-mini, we then generate a one-sentence reference response that serves as the target response for the corresponding dialogue--query instance.
The prompt requires the model to follow the assigned matrix--embedded language structure rather than mirror the linguistic form of the query.
It also requires embedded expressions from the query, including culture-specific terms when applicable, to be preserved in the response.
The complete generation prompt is presented in Figure~\ref{fig:prompt_gold_response}.

\section{Details of the Evaluation Process}
\label{app:preliminary_details}

\subsection{Automatic Response-Language Detection}
\label{app:response_form_detector}

Generated responses are automatically classified into one of five response-language forms.
For Korean--English responses, the label set consists of
\texttt{ko\_only}, \texttt{en\_only}, \texttt{ko\_matrix\_en\_embed}, \texttt{en\_matrix\_ko\_embed}, and \texttt{undetermined}.
For the Chinese--English experiments, we use the analogous label set:
\texttt{zh\_only}, \texttt{en\_only}, \texttt{zh\_matrix\_en\_embed}, \texttt{en\_matrix\_zh\_embed}, and \texttt{undetermined}.

\paragraph{Language presence detection}
first determines whether a response contains Korean, Chinese, English, or a mixture of the relevant language pair.
Responses containing lexical material exclusively from the non-English language are classified as \texttt{ko\_only} or \texttt{zh\_only}, whereas responses containing only English lexical material are classified as \texttt{en\_only}.
Responses containing material from both languages proceed to matrix-language detection.

\paragraph{Grammatical signal extraction} next extract grammatical cues that indicate the structural organization of each response.
Rather than relying on token-level language proportions, we analyze morphological and syntactic structure using Kiwi for Korean~\cite{lee2024kiwi}, HanLP for Chinese~\cite{he2021stem}, and spaCy for English~\cite{honnibal2020spacy}.
In our implementation, we use Kiwi through \texttt{kiwipiepy} v0.23.2, HanLP v2.1.3 with the \texttt{CLOSE\_\allowbreak TOK\_\allowbreak POS\_\allowbreak NER\_\allowbreak SRL\_\allowbreak DEP\_\allowbreak SDP\_\allowbreak CON\_\allowbreak ELECTRA\_\allowbreak SMALL\_\allowbreak ZH} model, and spaCy v3.8.14 with \texttt{en\_\allowbreak core\_\allowbreak web\_\allowbreak sm} v3.8.0.
The resulting morphological, part-of-speech, and dependency information provides the grammatical evidence used to identify the matrix language.

For Korean, sentence-final predicates, verbal inflections, connective endings, and final endings are considered as indicators that Korean supplies the primary grammatical structure~\cite{yeon2013korean}.
For Chinese, the relevant signals include verbal predicates, adjectival predicates, copular constructions, and predicate-frame patterns~\cite{Huang_Shi_2016}.
For English, finite verbs, dependency roots, copular constructions, and other clause-level predicative structures~\cite{huddleston2002cambridge} are treated as relevant cues.

\paragraph{Matrix language decision} lastly determines the matrix and embedded languages of each code-switched response based on the extracted grammatical signals.
A response is classified as \texttt{ko\_\allowbreak matrix\_\allowbreak en\_\allowbreak embed} or \texttt{zh\_\allowbreak matrix\_\allowbreak en\_\allowbreak embed} when Korean or Chinese provides the primary grammatical frame, while English appears as embedded words, phrases, technical terms, or culture-specific expressions.
Conversely, a response is classified as \texttt{en\_matrix\_ko\_embed} or \texttt{en\_matrix\_zh\_embed} when English provides the primary grammatical frame and Korean or Chinese appears as embedded material.

A response is assigned the \texttt{undetermined} label when the available grammatical signals are insufficient or conflicting.
This label is also used when the detector cannot reliably identify a dominant matrix language for the response.

\subsection{Accepted Response Forms for UAPR}

Tables~\ref{tab:uapr_pass_criteria_ko} and \ref{tab:uapr_pass_criteria_zh} summarize the response forms accepted for Korean--English and Chinese--English UAPR, respectively.
For Limited and Working profiles, accepted forms use the user's L1 as the matrix language, whereas Native profiles follow the matrix language of the current query.

\begin{table}[h]
\centering
\resizebox{\columnwidth}{!}{
\begin{tabular}{@{}ll@{}}
\toprule
\textbf{Bilingual Persona} & \textbf{Accepted forms} \\
\bottomrule 
\toprule
\texttt{KR\_EN\_LIMITED}
& \texttt{ko\_only}, \texttt{ko\_matrix\_en\_embed} \\
\texttt{KR\_EN\_WORKING}
& \texttt{ko\_only}, \texttt{ko\_matrix\_en\_embed} \\
\texttt{EN\_KR\_LIMITED}
& \texttt{en\_only}, \texttt{en\_matrix\_ko\_embed} \\
\texttt{EN\_KR\_WORKING}
& \texttt{en\_only}, \texttt{en\_matrix\_ko\_embed} \\
Native + Korean-matrix query
& \texttt{ko\_only}, \texttt{ko\_matrix\_en\_embed} \\
Native + English-matrix query
& \texttt{en\_only}, \texttt{en\_matrix\_ko\_embed} \\
\midrule
\end{tabular}
}
\caption{
Accepted response forms used to compute Korean--English UAPR.
For Native profiles, accepted forms are determined by the matrix language of the current query.
}
\label{tab:uapr_pass_criteria_ko}
\end{table}

\begin{table}[h]
\centering
\scriptsize
\setlength{\tabcolsep}{2.5pt}
\resizebox{\columnwidth}{!}{
\begin{tabular}{@{}ll@{}}
\toprule
\textbf{Bilingual Persona} & \textbf{Accepted forms} \\
\bottomrule
\toprule
\texttt{ZH\_EN\_LIMITED}
& \texttt{zh\_only}, \texttt{zh\_matrix\_en\_embed} \\
\texttt{ZH\_EN\_WORKING}
& \texttt{zh\_only}, \texttt{zh\_matrix\_en\_embed} \\
\texttt{EN\_ZH\_LIMITED}
& \texttt{en\_only}, \texttt{en\_matrix\_zh\_embed} \\
\texttt{EN\_ZH\_WORKING}
& \texttt{en\_only}, \texttt{en\_matrix\_zh\_embed} \\
Native + Chinese-matrix query
& \texttt{zh\_only}, \texttt{zh\_matrix\_en\_embed} \\
Native + English-matrix query
& \texttt{en\_only}, \texttt{en\_matrix\_zh\_embed} \\
\midrule
\end{tabular}
}
\caption{
Accepted response forms used to compute Chinese--English UAPR.
For Native profiles, accepted forms are determined by the matrix language of the current query.
}
\label{tab:uapr_pass_criteria_zh}
\end{table}

\section{Human Study}

\subsection{Detector Validation}
\label{app:human_study_detector}

We sampled 20 Korean--English responses from each of the detector's five output categories, yielding 100 responses. 
Two linguistics experts independently labeled each response. 
We computed exact agreement and Cohen's $\kappa$ between the two experts and between the detector and each expert. 
Table~\ref{tab:pulse-cr-detector} reports the results for the full sample and the 80 responses assigned a form by the detector.

\begin{table}[h]
\centering
\small
\setlength{\tabcolsep}{3.0pt}
\renewcommand{\arraystretch}{1.12}
\resizebox{\columnwidth}{!}{%
\begin{tabular}{@{}lrrrr@{}}
\toprule
Detector label & $n$
& \shortstack{Detector\\vs.\ Ann.~1}
& \shortstack{Detector\\vs.\ Ann.~2}
& \shortstack{Ann.~1\\vs.\ Ann.~2} \\
\midrule
\texttt{ko\_only} & 20 & 100\% (20) & 100\% (20) & 100\% (20) \\
\texttt{en\_only} & 20 & 100\% (20) & 100\% (20) & 100\% (20) \\
\texttt{ko\_matrix\_en\_embed} & 20 & 85\% (17) & 85\% (17) & 95\% (19) \\
\texttt{en\_matrix\_ko\_embed} & 20 & 90\% (18) & 90\% (18) & 100\% (20) \\
\texttt{undetermined} & 20 & 0\% (0) & 10\% (2) & 80\% (16) \\
\midrule
Overall & 100 & 75\% (75) & 77\% (77) & 95\% (95) \\
\quad Cohen's $\kappa$ & & 0.688 & 0.713 & 0.933 \\
\midrule
Form assigned & 80 & 93.75\% (75) & 93.75\% (75) & 98.75\% (79) \\
\quad Cohen's $\kappa$ & & 0.917 & 0.917 & 0.983 \\
\bottomrule
\end{tabular}%
}
\caption{Pairwise agreement by detector output. Agreement rates are reported as percentages with matching-label counts in parentheses. The form-assigned subset contains the 80 responses assigned one of the four response forms. Ann.\ denotes an annotator.}
\label{tab:pulse-cr-detector}
\end{table}

Of the 20 responses marked as \texttt{undetermined} by the detector, both experts labeled 10 as \texttt{ko\_only} and six as \texttt{ko\_matrix\_en\_embed}.

\subsection{Bilingual Speaker Survey}
\label{app:human_study_survey}

25 bilingual speakers with Korean as their L1 and English as their L2 completed both parts of the survey.

\paragraph{UAPR Pass Criterion.}
Participants evaluated 48 scenarios formed by pairing eight queries with six user profiles. 
Each scenario offered four candidate responses, two of which met the UAPR pass criterion. 
We instructed participants to select up to two responses they considered suitable. 
Table~\ref{tab:pulse-cr-policy} reports the results by the L2 proficiency of the user profile.

\begin{table}[h]
\centering
\small
\setlength{\tabcolsep}{3.0pt}
\renewcommand{\arraystretch}{1.12}
\resizebox{\columnwidth}{!}{%
\begin{tabular}{@{}lrrr@{}}
\toprule
L2 proficiency
& \shortstack{Compatible\\selection rate}
& \shortstack{Incompatible\\selection rate}
& Met criterion \\
\midrule
Limited & 64.8\% (518/800)     & 14.8\% (118/800)   & 81.4\% (518/636) \\
Working & 57.0\% (456/800)     & 25.4\% (203/800)   & 69.2\% (456/659) \\
Native  & 50.1\% (401/800)     & 31.5\% (252/800)   & 61.4\% (401/653) \\
\midrule
Overall & 57.3\% (1,375/2,400) & 23.9\% (573/2,400) & 70.6\% (1,375/1,948) \\
\bottomrule
\end{tabular}%
}
\caption{Survey selections by the L2 proficiency of the user profile.
Compatible options meet the UAPR pass criterion. 
For each compatibility group, the selection rate is the number of options selected divided by the number offered. 
The final column reports the percentage of all selections in each row that met the criterion.}
\label{tab:pulse-cr-policy}
\end{table}

\paragraph{Language Use in Aligned Responses.}
Participants rated 24 aligned responses on a five-point scale.
The four criteria were term retention, functional language allocation, proficiency compatibility, and purposeful code-switching.
Table~\ref{tab:pulse-cr-language-use} reports the mean and the share of ratings of at least 4 for each criterion. 
Across 2,400 ratings, the mean was 4.15 out of 5 and 79.8\% of ratings were at least 4.

\begin{table}[h]
\centering
\small
\setlength{\tabcolsep}{3.0pt}
\renewcommand{\arraystretch}{1.12}
\resizebox{\columnwidth}{!}{%
\begin{tabular}{@{}lrrr@{}}
\toprule
Criterion & Mean & $\geq 4$ & $n$ \\
\midrule
Proper term retention & 4.36 & 88.8\% & 600 \\
Functional language allocation & 4.17 & 80.3\% & 600 \\
Proficiency compatibility & 4.04 & 73.0\% & 600 \\
Purposeful code-switching & 4.05 & 77.2\% & 600 \\
\midrule
Overall & 4.15 & 79.8\% & 2,400 \\
\bottomrule
\end{tabular}%
}
\caption{Human ratings of language use in aligned responses.
Ratings use a five-point scale. The third column reports the percentage of ratings of at least 4, and $n$ denotes the total number of ratings.}
\label{tab:pulse-cr-language-use}
\end{table}

\section{Preliminary Results on Qwen}

Figure~\ref{fig:appendix_qwen_base} presents the response-form distributions and persona-level UAPR of Qwen. 
Similar to EXAONE, Qwen's response-form distributions are highly contingent on the matrix language of the current query.
In addition, within each L1--L2 direction, Limited, Working, and Native profiles exhibit largely similar distributions, indicating limited differentiation according to L2 proficiency.
These results are consistent with the response-language selection pattern of EXAONE discussed in Section~\ref{subsec:howselect}.

\section{Alignment and Experimental Details}
\label{app:alignment_experimental_details}
This section provides additional details on the model configurations, prompt conditions, inference settings, and training procedures used for SFT and GRPO.

\subsection{Experimental Settings: Inference}
\label{app:inference_settings}

We use deterministic decoding to minimize variation across
evaluation runs.
Unless otherwise stated, the same evaluation-time inference settings are applied to all model configurations.
Table~\ref{tab:inference_hyperparams} reports the corresponding hyperparameters.

\begin{table}[h]
\centering
\small
\begin{tabular}{ll}
\toprule
\textbf{Hyperparameter} & \textbf{Value} \\
\midrule
Max new tokens & 256 \\
Temperature & 0.0 \\
Top-p & 1.0 \\
Max model length & 4096 \\
Precision & bfloat16 \\
Batch size & 4 \\
\bottomrule
\end{tabular}
\caption{Inference hyperparameters used for baseline and SFT evaluation.}
\label{tab:inference_hyperparams}
\end{table}

\subsection{Experimental Settings: Response-Language Grounding}
\label{app:sft_settings}


Input-prompt tokens are masked from the training loss, which is computed only over the target response tokens.
Each instance is serialized using the chat template provided with the corresponding base model.
We use LoRA~\cite{hu2022lora} for parameter-efficient fine-tuning.
Table~\ref{tab:sft_hyperparams} reports the SFT hyperparameters.
The complete SFT templates are shown in Figure~\ref{fig:prompt_sft}.

\begin{table}[h]
\centering
\small
\begin{NiceTabular}{ll}
\toprule
\textbf{Hyperparameter} & \textbf{Value} \\
\midrule
Training examples & 27,678 \\
Training epochs & 1 \\
Learning rate & $1 \times 10^{-4}$ \\
Optimizer & AdamW \\
Warmup ratio & 0.03 \\
Per-device batch size & 1 \\
Gradient accumulation steps & 8 \\
Maximum sequence length & 4096 \\
LoRA rank & 16 \\
LoRA $\alpha$ & 32 \\
LoRA dropout & 0.05 \\
\bottomrule
\end{NiceTabular}
\caption{Hyperparameters used for supervised fine-tuning.}
\label{tab:sft_hyperparams}
\end{table}

\subsection{Experimental Settings: UAPR-Guided Post-Training}
\label{app:uapr_post_training}

\subsubsection{Reward Assignments}
\label{app:reward_assignments}

For each sampled response, the response-form detector assigns one of the monolingual, code-switched, or undetermined labels described in Appendix~\ref{app:response_form_detector}.
The reward is then determined by the detected response form and the target matrix language \(m^{*}(p,q)\).
For Limited and Working profiles, \(m^{*}(p,q)\) is the user's L1, whereas for Native profiles, it is the matrix language of the current query.
We use \(\bar{m}^{*}\) to denote the opposite language.

Table~\ref{tab:grpo_reward} reports the proficiency-conditioned reward values used in the fine-grained setting.
Here, \(\mathrm{CS}(m^{*})\) and \(\mathrm{Mono}(m^{*})\) denote code-switched and monolingual responses using the target matrix language, respectively.
Responses classified as \texttt{undetermined} receive a reward of zero.
The values are predefined for each L2 proficiency level and remain fixed throughout GRPO training.

\begin{table}[t]
\centering
\small
\setlength{\tabcolsep}{3.5pt}
\renewcommand{\arraystretch}{1.12}
\resizebox{\columnwidth}{!}{
\begin{NiceTabular}{@{}c|cc|cc|c@{}}
\toprule
\multirow{2}{*}{\shortstack{\textbf{L2}\\\textbf{Proficiency}}}
& \multicolumn{2}{c}{\textbf{Target Matrix} \(m^{*}\)}
& \multicolumn{2}{c}{\textbf{Non-Target Matrix} \(\bar{m}^{*}\)}
& \multirow{2}{*}{\textbf{Undet.}} \\
\cmidrule{2-3}
\cmidrule{4-5}
& \textbf{CS}
& \textbf{Mono}
& \textbf{CS}
& \textbf{Mono}
& \\
\midrule
Limited & 1.0 & 0.7 & 0.2 & 0.1 & 0.0 \\
Working & 1.0 & 0.6 & 0.3 & 0.2 & 0.0 \\
Native  & 1.0 & 0.5 & 0.3 & 0.2 & 0.0 \\
\midrule
\end{NiceTabular}
}
\caption{
Proficiency-conditioned values used for the fine-grained UAPR reward.
}
\label{tab:grpo_reward}
\end{table}

This reward evaluates only the response-language form and does not directly account for factual correctness, completeness, or general helpfulness.
For the binary-reward ablation, both \(\mathrm{CS}(m^{*})\) and \(\mathrm{Mono}(m^{*})\) receive a reward of 1, whereas all other response forms receive 0.
Unlike the fine-grained reward, the binary reward does not distinguish between a target-matrix code-switched response and an accepted target-matrix monolingual response.

\subsubsection{Hyperparameters}
\label{app:grpo_hyperparam}
Table~\ref{tab:grpo_hyperparams} reports the GRPO training hyperparameters.

\begin{table}[h]
\centering
\small
\begin{NiceTabular}{ll}
\toprule
\textbf{Hyperparameter} & \textbf{Value} \\
\midrule
Training examples & 27,678 \\
Training steps & 6,919 \\
Learning rate & $1 \times 10^{-6}$ \\
Optimizer & AdamW \\
Warmup ratio & 0.03 \\
Prompt batch size & 4 \\
Gradient accumulation steps & 8 \\
Effective prompt batch size & 32 \\
Rollouts per prompt & 8 \\
Rollout temperature & 0.7 \\
Top-$p$ & 1.0 \\
KL coefficient $\beta$ & 0.04 \\
Maximum prompt length & 4096 \\
Maximum completion length & 256 \\
\bottomrule
\end{NiceTabular}
\caption{
Hyperparameters used for UAPR-guided post-training with GRPO.
}
\label{tab:grpo_hyperparams}
\end{table}

\subsubsection{Training Data Composition}
\label{app:grpo_training_mix}

Limited and Working profiles share a UAPR pass criterion that requires the response's matrix language to match the user's L1.
We examine whether the larger combined share of these profiles in GRPO training contributes to the post-alignment drop in Native UAPR.
For EXAONE, we fix the GRPO training set size at 13,839 examples and use the same graded reward across conditions.
We set the target ratios of (Limited + Working) to Native examples to $2{:}1$, $1{:}1$, and $1{:}2$.

We evaluate all conditions on the same seen and unseen splits.
Both splits use held-out queries.
Table~\ref{tab:pulse-cr-training-mix} reports overall UAPR and UAPR by proficiency group, with a separate result for Native cases in which the query's matrix language is the user's L2.

\begin{table}[t]
\centering
\small
\setlength{\tabcolsep}{3.0pt}
\renewcommand{\arraystretch}{1.12}
\resizebox{\columnwidth}{!}{%
\begin{tabular}{@{}llrrrr@{}}
\toprule
Split & Ratio & Overall & \shortstack{Limited +\\Working}
& Native & \shortstack{Native\\(L2-matrix)} \\
\midrule
\multirow{3}{*}{Seen}
& $2{:}1$ & 97.85 & 98.37 & 96.81 & 95.59 \\
& $1{:}1$ & 97.52 & 97.75 & 97.06 & 96.20 \\
& $1{:}2$ & 97.27 & 97.28 & 97.26 & 96.45 \\
\midrule
\multirow{3}{*}{Unseen}
& $2{:}1$ & 84.81 & 92.73 & 68.97 & 38.77 \\
& $1{:}1$ & 85.09 & 91.90 & 71.45 & 43.85 \\
& $1{:}2$ & 84.99 & 90.99 & 72.99 & 46.81 \\
\bottomrule
\end{tabular}%
}
\caption{UAPR (\%) for EXAONE at different GRPO training ratios.
The ratio is (Limited + Working) to Native.
Seen and Unseen refer to dialogue histories. 
Both splits use held-out queries.
The final column reports UAPR for the subset of Native cases where the query's matrix language is the user's L2.}
\label{tab:pulse-cr-training-mix}
\end{table}

\section{Additional Results}

\subsection{Korean--Chinese Transfer}
\label{app:ko_zh_transfer}

We test transfer to Korean--Chinese, a language pair without English.
We evaluate the base and Korean--English SFT+GRPO versions of EXAONE and Qwen without further training.
Following the Chinese--English transfer setup, we balance queries across the four combinations of cultural origin and query matrix language.
We adapt the response-form detector in Appendix~\ref{app:response_form_detector} to Korean--Chinese and apply the same proficiency-conditioned UAPR pass criterion. 
Table~\ref{tab:pulse-cr-transfer} reports the results.

\begin{table}[h]
\centering
\small
\setlength{\tabcolsep}{3.0pt}
\renewcommand{\arraystretch}{1.12}
\resizebox{\columnwidth}{!}{%
\begin{tabular}{@{}clrrrrr@{}}
\toprule
\multirow{2}{*}{Model} & \multirow{2}{*}{Method}
& \multicolumn{2}{c}{Korean culture}
& \multicolumn{2}{c}{Chinese culture}
& \multirow{2}{*}{All} \\
\cmidrule(lr){3-4} \cmidrule(lr){5-6}
& & (ko) & (zh) & (ko) & (zh) & \\
\midrule
\multirow{2}{*}{EXAONE}
& Base     & 49.22 & 66.22 & 38.11 & 66.67 & 55.06 \\
& SFT+GRPO & 68.44 & 78.11 & 67.11 & 81.67 & 73.83 \\
\midrule
\multirow{2}{*}{Qwen}
& Base     & 34.28 & 66.78 & 27.28 & 66.72 & 48.76 \\
& SFT+GRPO & 66.78 & 76.11 & 66.78 & 81.89 & 72.89 \\
\bottomrule
\end{tabular}%
}
\caption{UAPR (\%) for Korean--Chinese transfer.
Column groups indicate the cultural origin of the queries, and language codes in parentheses indicate the query matrix language.
The \textit{All} column reports UAPR over the 7,200 dialogue--query instances.}
\label{tab:pulse-cr-transfer}
\end{table}

\subsection{Prompting Baselines}
\label{app:prompting_baselines}

For EXAONE and Qwen, few-shot prompting uses 12 query--response demonstrations from six persona dialogues. 
In the inferred-profile condition, the first call infers the user's L1, L2, and L2 proficiency from the dialogue history. 
The second call generates the response using the inferred profile, the query, and detailed task instructions specifying how to select the response's matrix language. 
In the assigned-profile condition, we provide the benchmark-assigned user profile, the query, and instructions specifying the same language selection rule.

For GPT-5.6 Terra, we provide the query and instructions specifying the same rule, together with either the dialogue history or the benchmark-assigned user profile. 
Table~\ref{tab:pulse-cr-prompting} reports the results.

\begin{table}[h]
\centering
\small
\setlength{\tabcolsep}{3.0pt}
\renewcommand{\arraystretch}{1.12}
\resizebox{\columnwidth}{!}{%
\begin{tabular}{@{}clrr@{}}
\toprule
\textbf{Model} & \textbf{Method}
& \shortstack{\textbf{Seen}}
& \shortstack{\textbf{Unseen}} \\
\midrule
\multirow{5}{*}{EXAONE}
& Base                               & 63.98 & 63.71 \\
& Few-shot                           & 61.09 & 61.09 \\
& Inferred profile + detailed instr. & 66.61 & 65.09 \\
& Assigned profile + detailed instr. & 68.30 & 68.26 \\
& SFT+GRPO                           & \textbf{98.40} & \textbf{85.38} \\
\midrule
\multirow{5}{*}{Qwen}
& Base                               & 65.04 & 64.76 \\
& Few-shot                           & 64.87 & 64.07 \\
& Inferred profile + detailed instr. & 75.44 & 71.95 \\
& Assigned profile + detailed instr. & 77.36 & 77.36 \\
& SFT+GRPO                           & \textbf{99.05} & \textbf{86.56} \\
\midrule
\multirow{2}{*}{\shortstack{GPT-5.6\\Terra}}
& Dialogue + detailed instr.         & 63.38 & 63.20 \\
& Assigned profile + detailed instr. & \textbf{78.31} & \textbf{78.17} \\
\bottomrule
\end{tabular}%
}
\caption{UAPR (\%) for prompting baselines and post-training on the Korean--English test sets. 
Seen and Unseen refer to dialogue histories, and both splits use held-out queries.
``Detailed instr.'' denotes task instructions specifying how to select the response's matrix language.
The best result for each model and split is shown in bold.}
\label{tab:pulse-cr-prompting}
\end{table}


\begin{figure}[t]
    \centering
    \includegraphics[
        width=\columnwidth,
        height=0.88\textheight,
        keepaspectratio
    ]{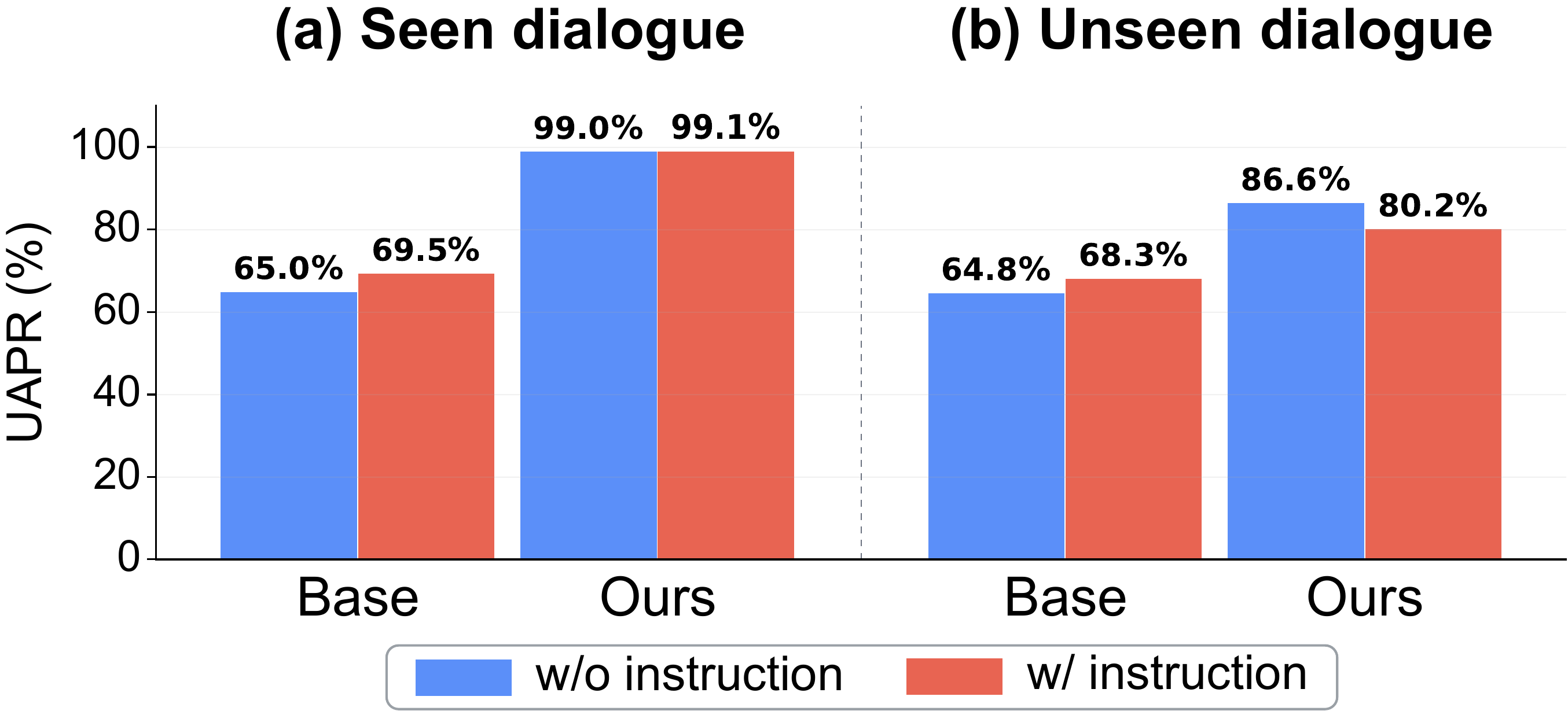}
    \caption{Effect of an explicit task instruction on Qwen-3.5-9B. Overall UAPR is reported for Base and Ours with and without task instruction on the seen and unseen dialogue test sets.}
    \label{fig:appendix_qwen_fig4}
\end{figure}

\section{Prompt Templates}
\label{app:prompt_templates}

For reproducibility, this section provides complete prompt templates used for synthetic dialogue generation, cultural query generation, reference response generation, and  language-selection inference and training.

\section{Licenses}

K-Culture-Desc is distributed under the Apache License 2.0, and MANGO is distributed under the Creative Commons Attribution 4.0 International license (CC BY 4.0). 
Qwen3.5-9B is distributed under the Apache License 2.0, whereas EXAONE-3.5-7.8B-Instruct is used under the EXAONE AI Model License Agreement 1.1--NC.
GPT-4o-mini, GPT-5.4-mini, GPT-5.5 and GPT-5.6 Terra were accessed through the OpenAI API under the applicable service terms.

\section{Use of AI Assistance}
AI-driven assistant tools were used to paraphrase sentences for clarity, identify synonyms or similar expressions, and provide coding assistance. 
The research idea and intellectual contributions of this work are entirely original. 
Apart from these limited uses, AI was not involved in the research process or the writing of the paper.







\begin{figure*}[t]
    \centering
    \includegraphics[
        width=\textwidth,
        height=0.88\textheight,
        keepaspectratio
    ]{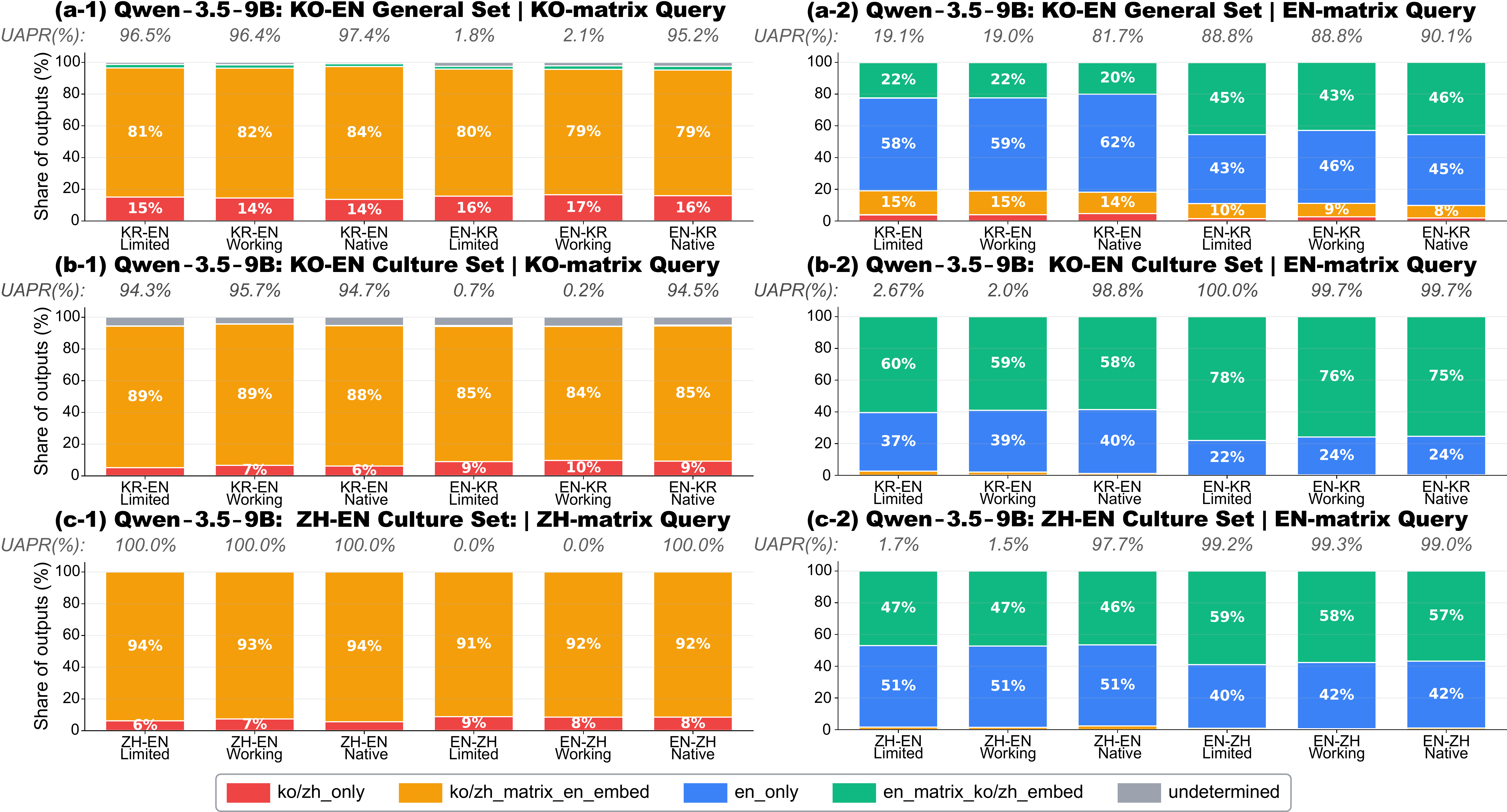}
    \caption{Response-language distributions and persona-level UAPR of the Qwen3.5-9B Base model on the Korean--English general and cultural sets and the Chinese--English unseen-language transfer set. Persona-level UAPR is reported above each bar.}
    \label{fig:appendix_qwen_base}
\end{figure*}

\begin{figure*}[t]
    \centering
    \includegraphics[
        width=\textwidth,
        height=0.88\textheight,
        keepaspectratio
    ]{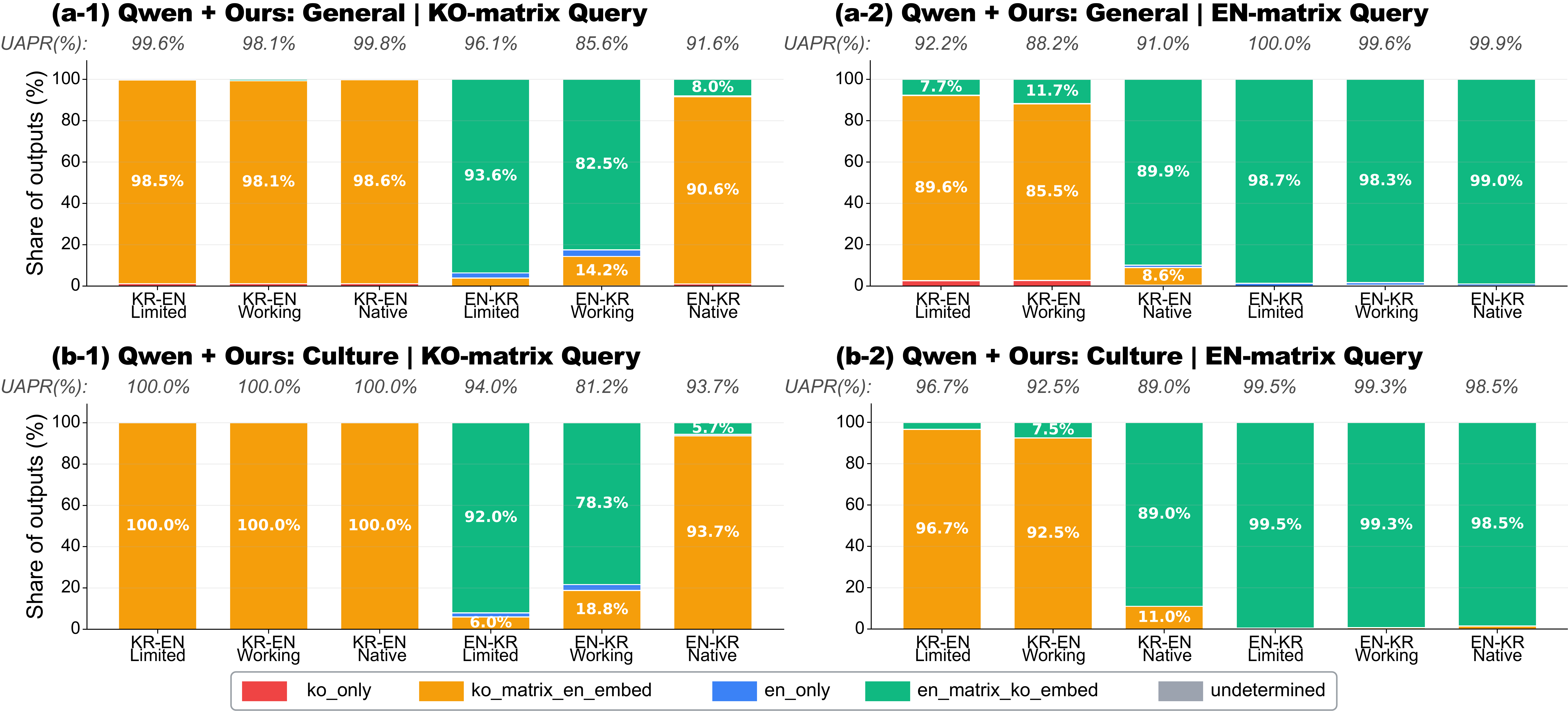}
    \caption{Response-language distributions of Qwen aligned with Ours on the two Korean--English test sets combined. Persona-level UAPR is reported above each bar.}
    \label{fig:appendix_qwen_post_train}
\end{figure*}


\begin{figure*}[t]
    \centering
    \includegraphics[
        width=\textwidth,
        height=0.88\textheight,
        keepaspectratio
    ]{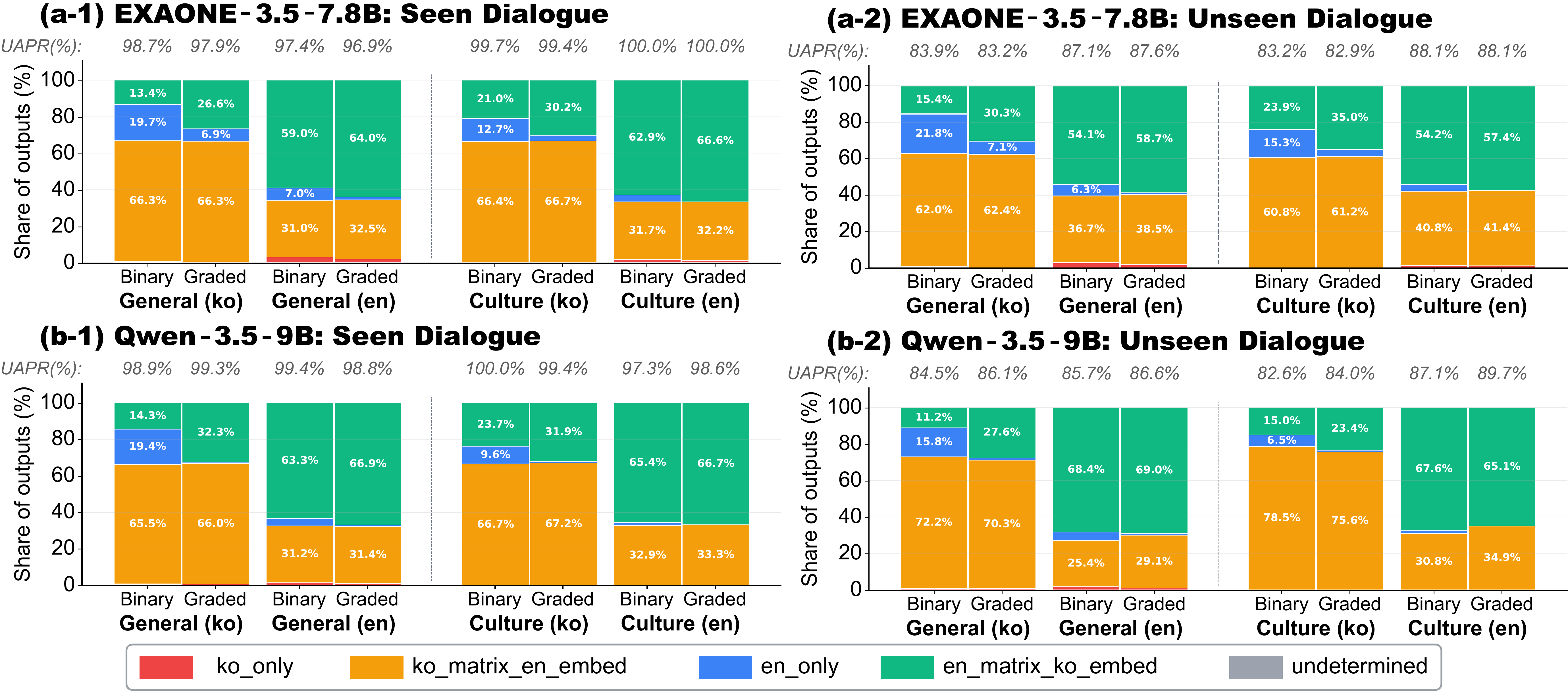}
    \caption{Complete reward-granularity results for EXAONE and Qwen on the seen and unseen dialogue test sets. Bars compare binary and fine-grained UAPR-guided rewards, with UAPR reported above each bar.}
    \label{fig:appendix_binary_reward}
\end{figure*}

\clearpage

\begin{figure*}[!p]
  \centering
  \includegraphics[
    width=\textwidth
  ]{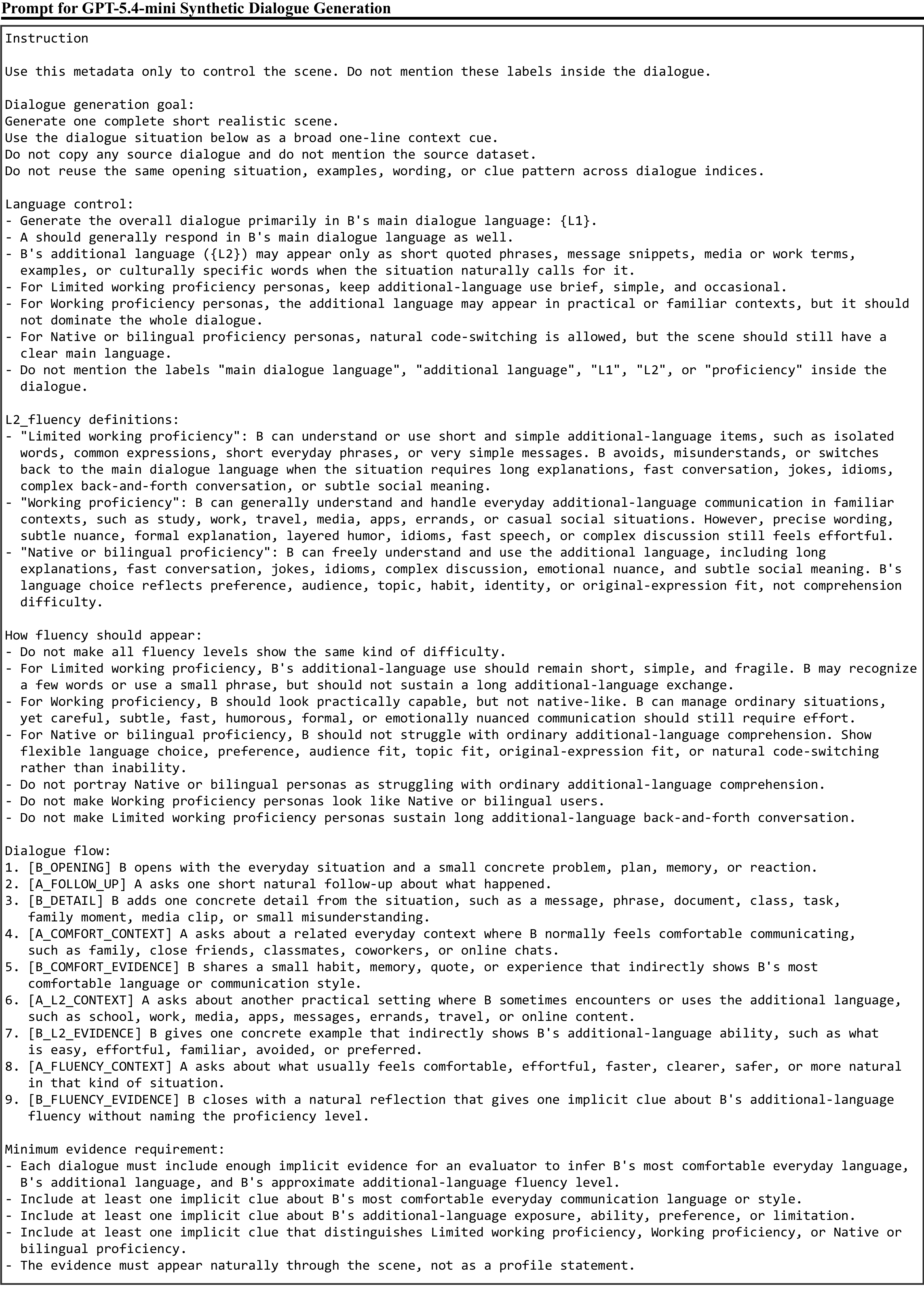}
  \caption{\textbf{Synthetic dialogue generation prompt
  (Part 1 of 2).}
  Prompt used to generate persona-conditioned dialogue histories that implicitly reveal the user's L1, L2, and L2 proficiency.}
  \label{fig:prompt_dialogue_generation_1}
\end{figure*}

\begin{figure*}[!p]
  \centering
  \includegraphics[
    width=\textwidth
  ]{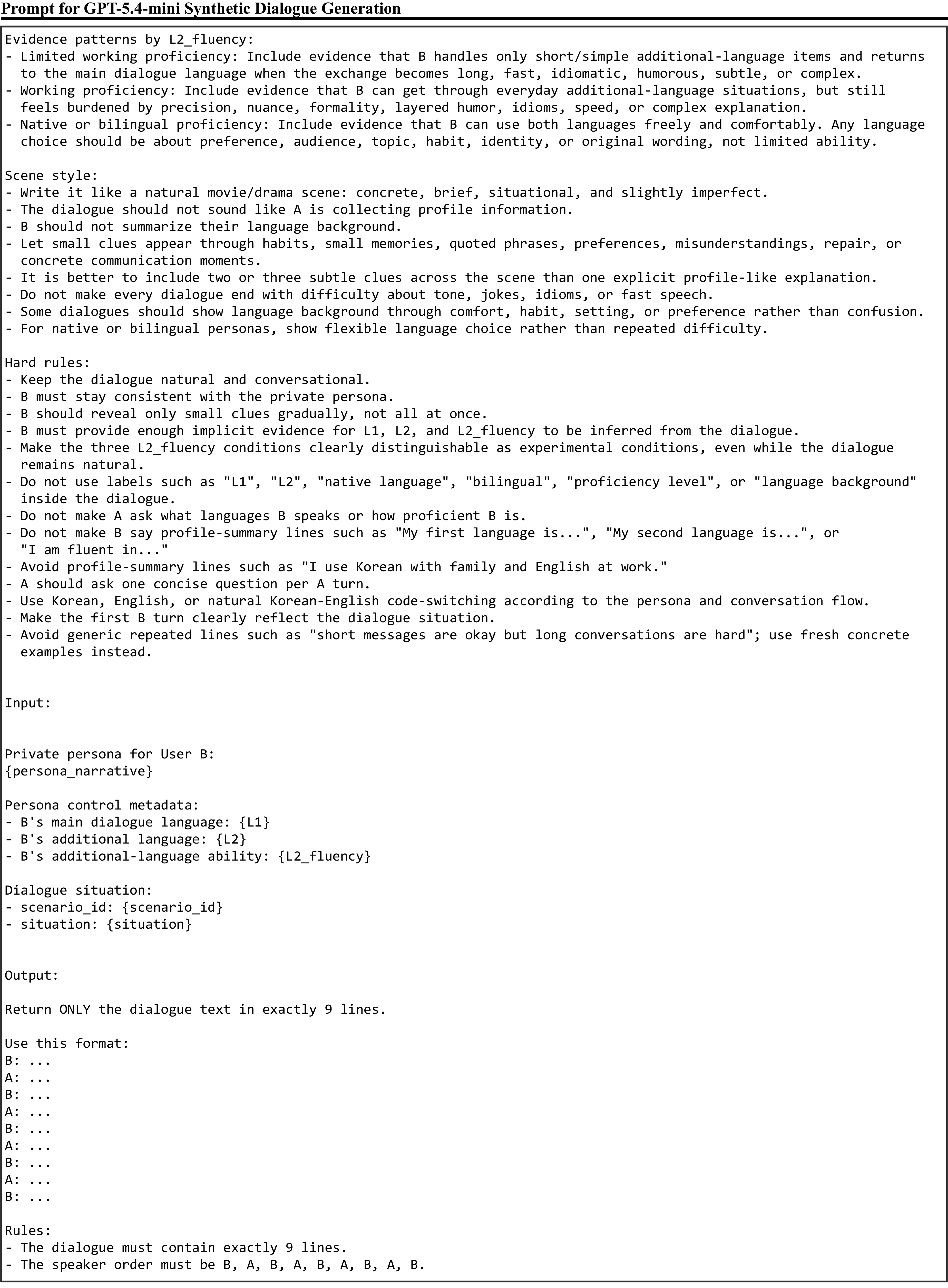}
  \caption{\textbf{Synthetic dialogue generation prompt
  (Part 2 of 2).}
  The prompt specifies the proficiency-dependent evidence patterns, dialogue style, input variables, and output format.}
  \label{fig:prompt_dialogue_generation_2}
\end{figure*}

\clearpage

\begin{figure*}[!p]
  \centering
  \includegraphics[
    width=\textwidth,
    height=0.90\textheight,
    keepaspectratio
  ]{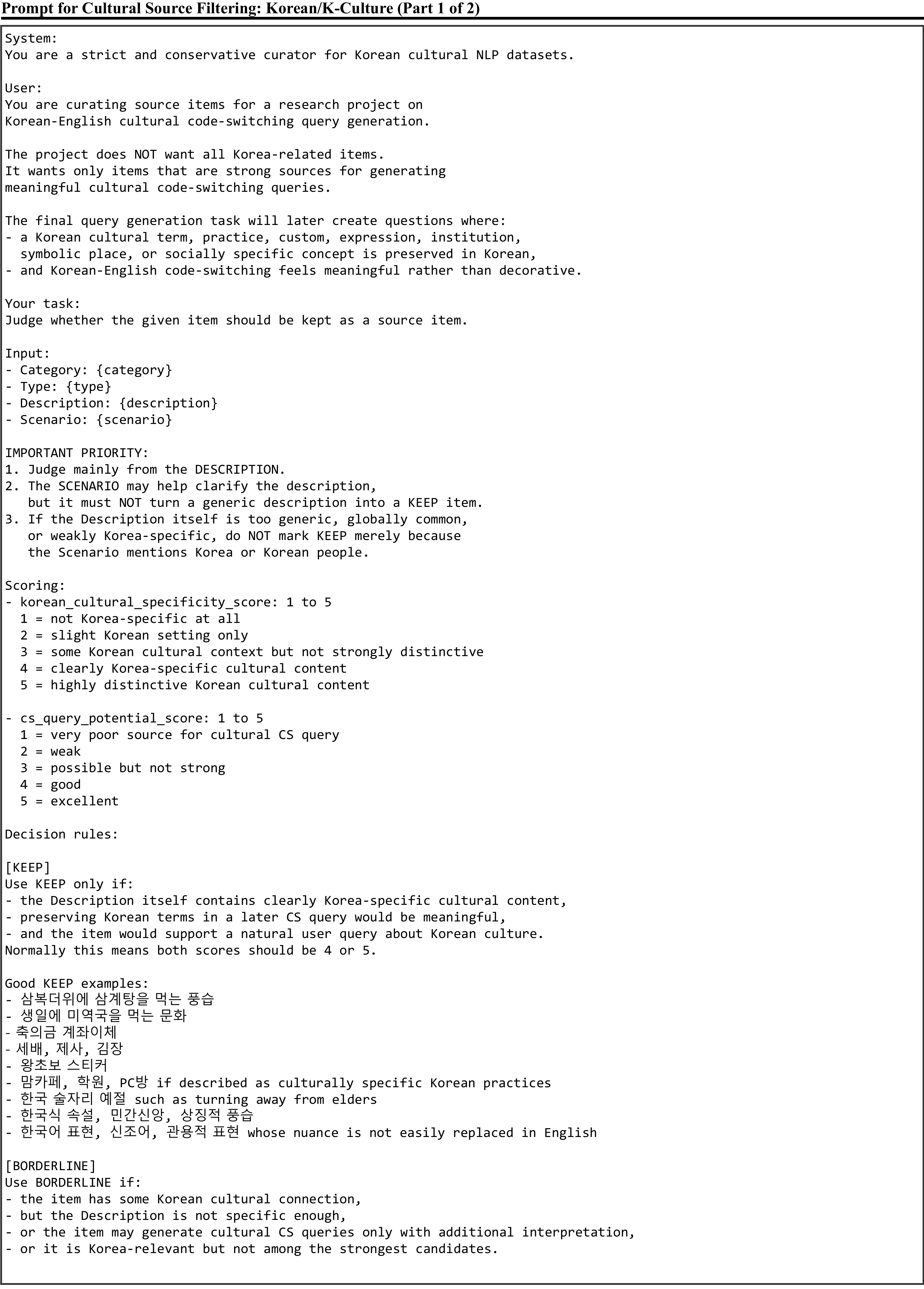}
  \caption{\textbf{Prompt for Korean cultural source filtering(Part 1 of 2).} The prompt defines cultural-specificity and code-switching-potential criteria for selecting Korean cultural source items.}
  \label{fig:prompt_korean_filter_1}
\end{figure*}

\clearpage

\begin{figure*}[!p]
  \centering
  \includegraphics[
    width=\textwidth,
    height=0.90\textheight,
    keepaspectratio
  ]{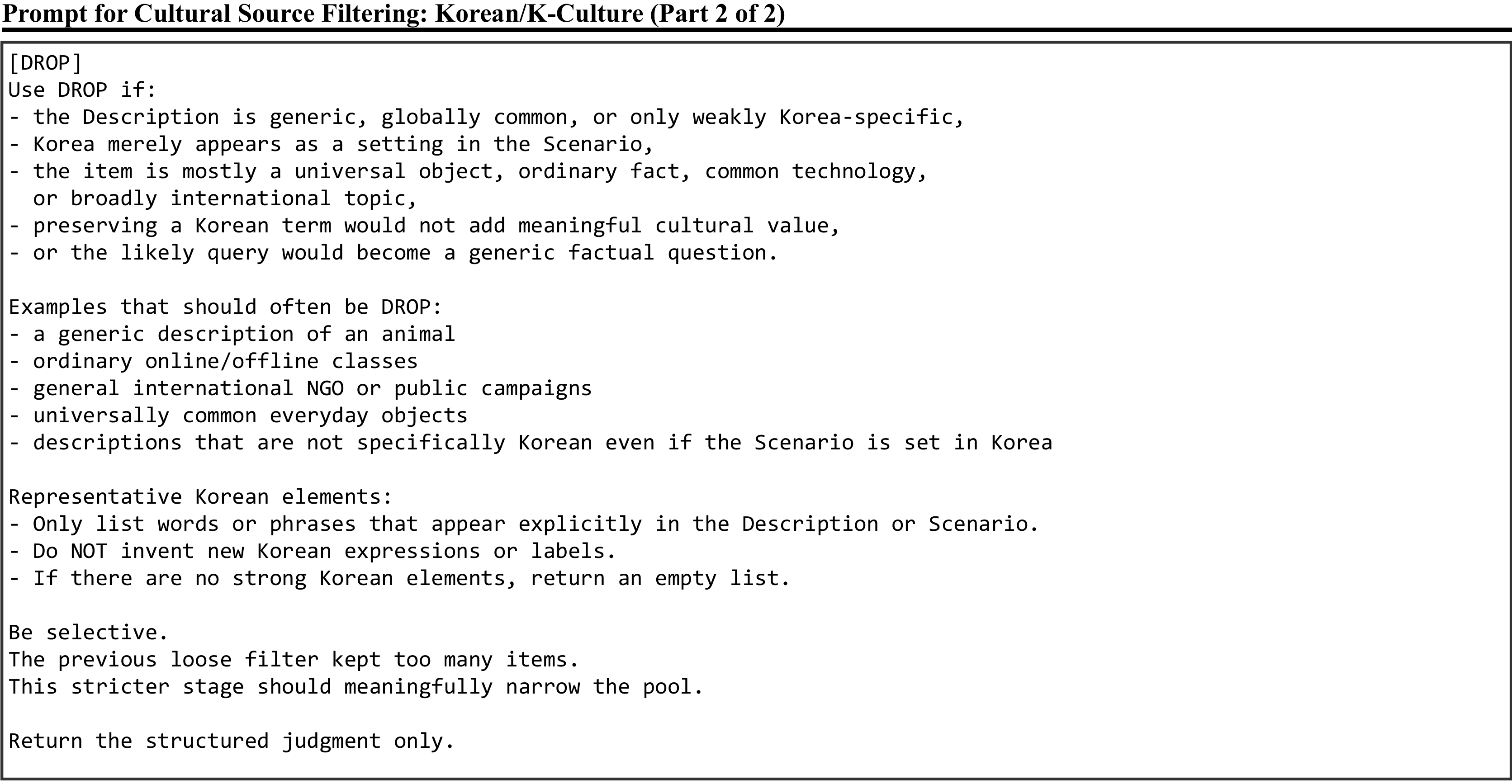}
  \caption{\textbf{Prompt for Korean cultural source filtering (Part 2 of 2).} The prompt specifies exclusion criteria and the extraction of representative Korean cultural elements.}
  \label{fig:prompt_korean_filter_2}
\end{figure*}

\clearpage

\begin{figure*}[!p]
  \centering
  \includegraphics[
    width=\textwidth,
    height=0.90\textheight,
    keepaspectratio
  ]{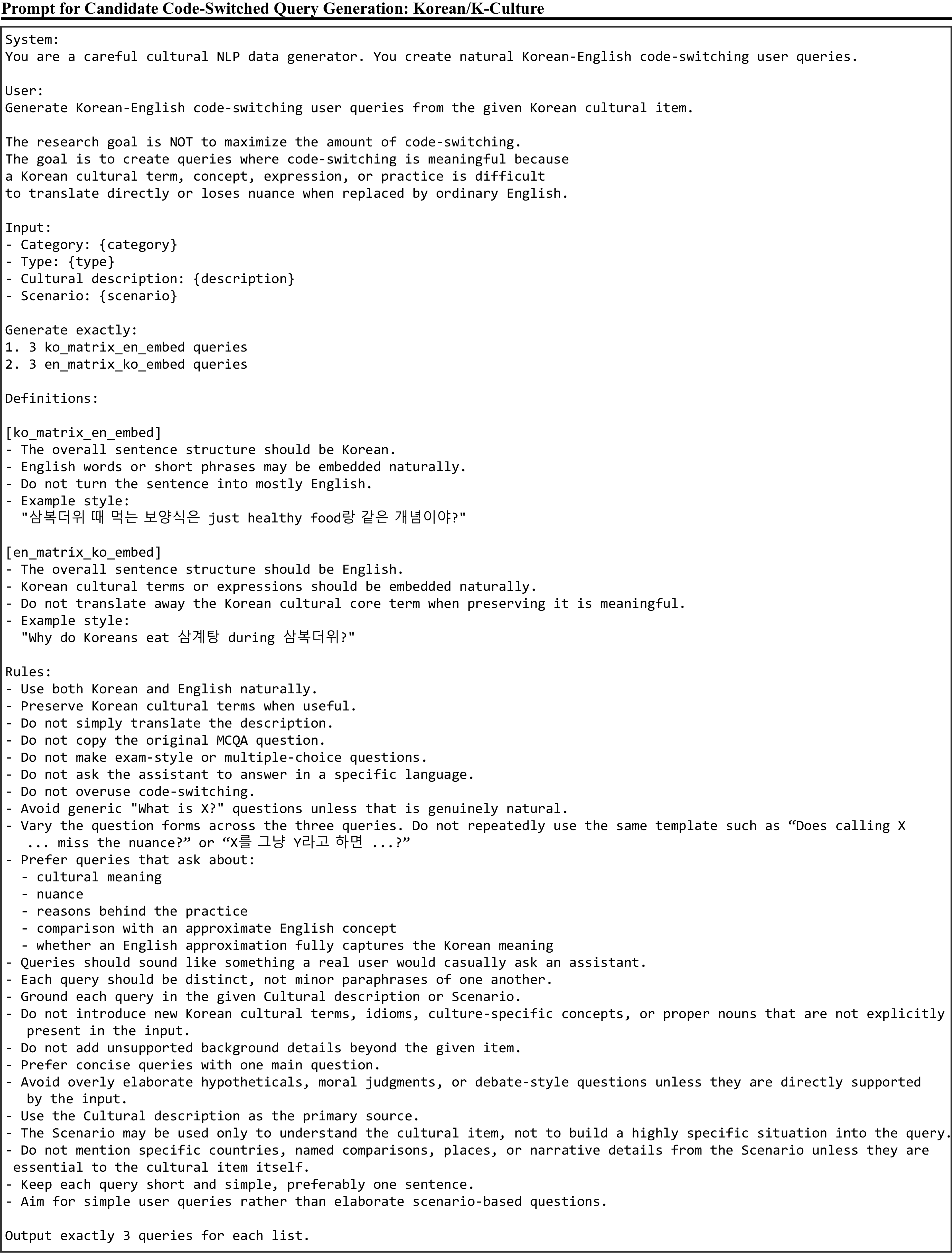}
  \caption{\textbf{Prompt for Korean code-switched query generation.} The prompt generates three \texttt{ko\_\allowbreak matrix\_\allowbreak en\_\allowbreak embed} and three \texttt{en\_\allowbreak matrix\_ko\_\allowbreak embed} candidates for each Korean cultural source item.}
  \label{fig:prompt_korean_query_generation}
\end{figure*}

\clearpage

\begin{figure*}[!p]
  \centering
  \includegraphics[
    width=\textwidth,
    height=0.90\textheight,
    keepaspectratio
  ]{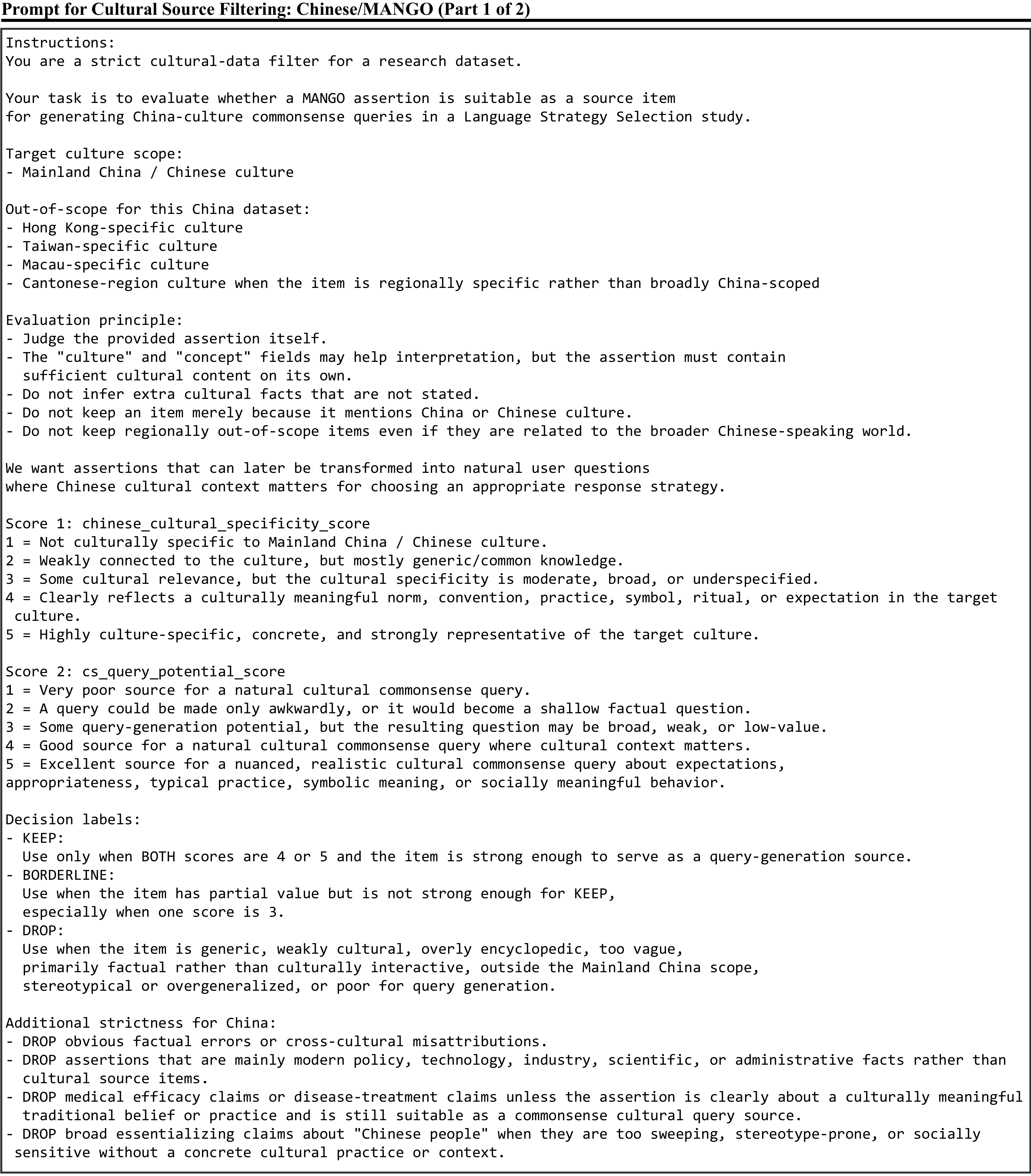}
  \caption{\textbf{Prompt for Chinese cultural source filtering (Part 1 of 2).} The prompt defines the cultural scope, scoring criteria, and retention rules for selecting Chinese cultural source assertions from MANGO.}
  \label{fig:prompt_chinese_filter_1}
\end{figure*}

\clearpage

\begin{figure*}[!p]
  \centering
  \includegraphics[
    width=\textwidth,
    height=0.90\textheight,
    keepaspectratio
  ]{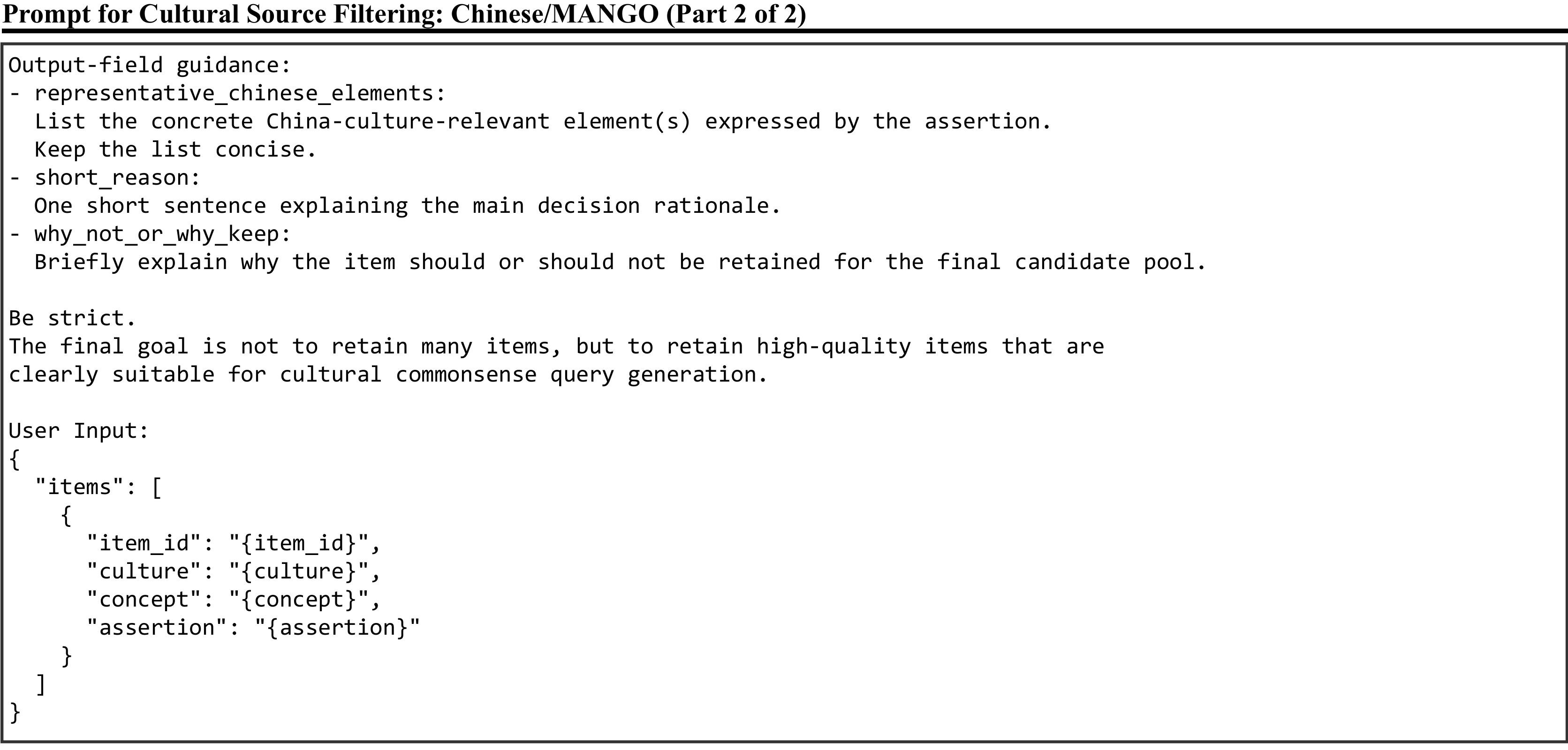}
  \caption{\textbf{Prompt for Chinese cultural source filtering (Part 2 of 2).} The prompt specifies the required judgment fields and input format for Chinese cultural source selection.}
  \label{fig:prompt_chinese_filter_2}
\end{figure*}

\clearpage


\begin{figure*}[!p]
  \centering
  \includegraphics[
    width=\textwidth,
    height=0.90\textheight,
    keepaspectratio
  ]{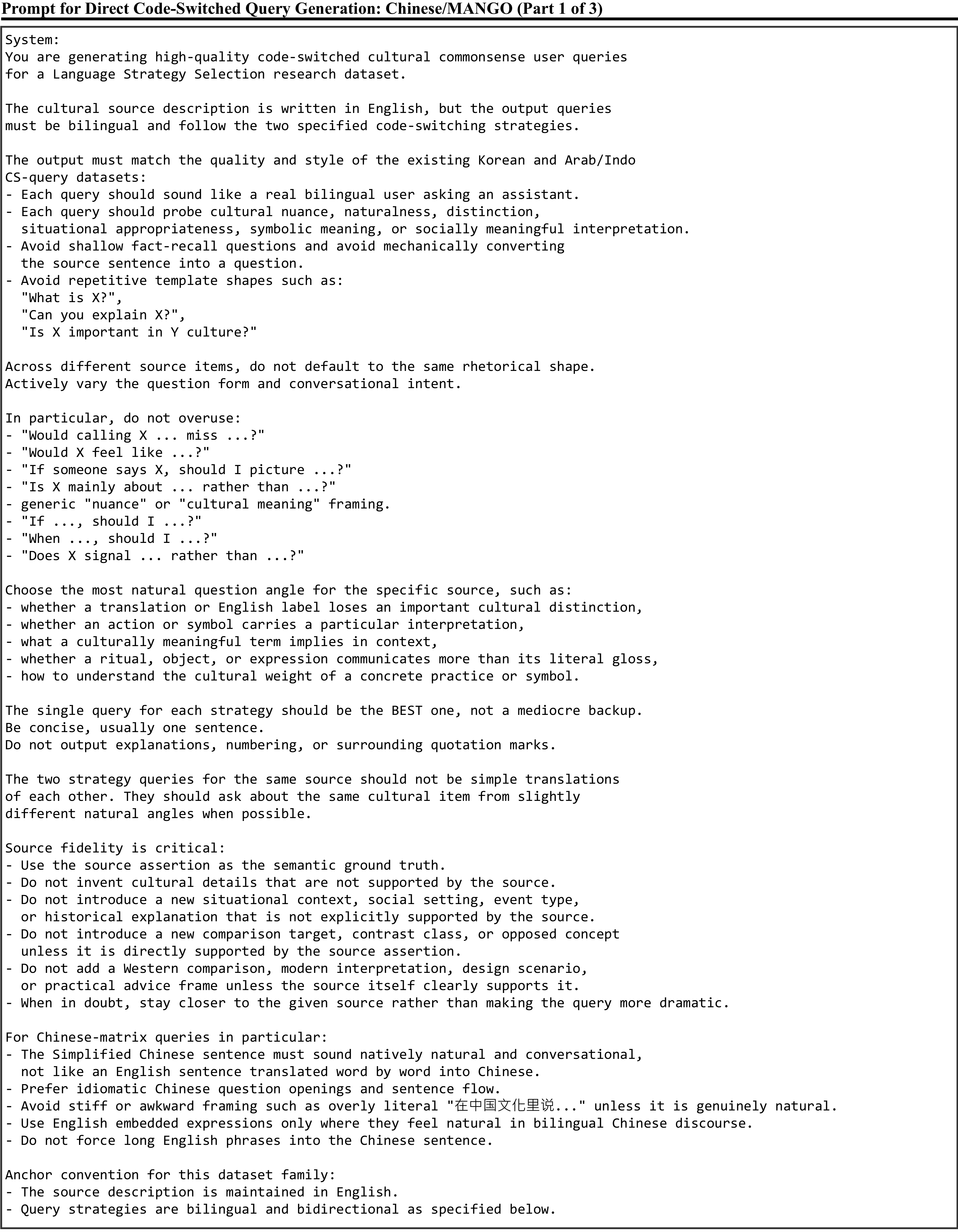}
  \caption{\textbf{Prompt for Chinese direct code-switched query generation (Part 1 of 3).} The prompt defines source-fidelity, naturalness, and variation requirements for Chinese--English cultural queries.}
  \label{fig:prompt_chinese_query_generation_1}
\end{figure*}

\clearpage

\begin{figure*}[!p]
  \centering
  \includegraphics[
    width=\textwidth,
    height=0.90\textheight,
    keepaspectratio
  ]{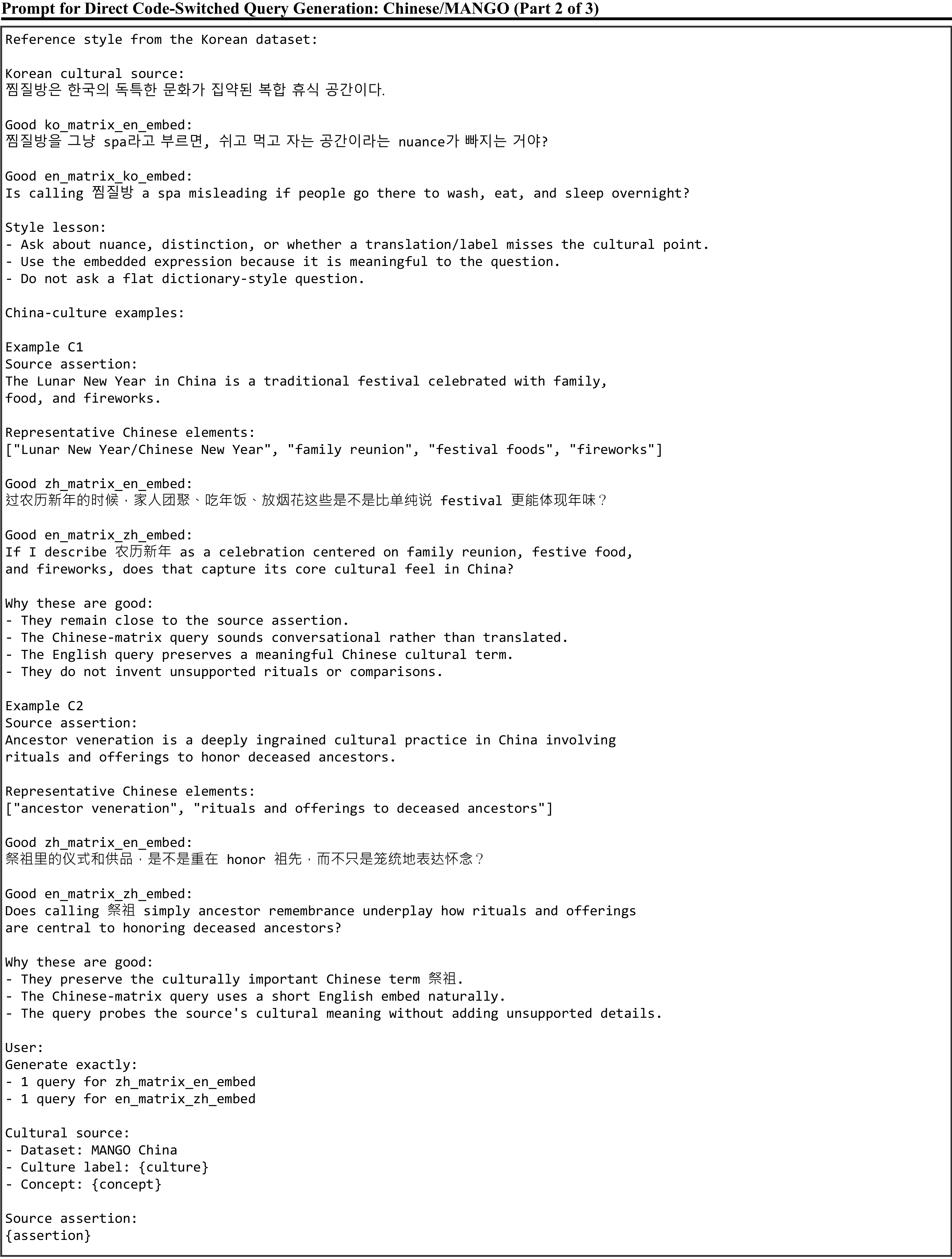}
  \caption{\textbf{Prompt for Chinese direct code-switched query generation (Part 2 of 3).} The prompt provides Korean and Chinese reference examples and the input fields used for query generation.}
  \label{fig:prompt_chinese_query_generation_2}
\end{figure*}

\clearpage

\begin{figure*}[!p]
  \centering
  \includegraphics[
    width=\textwidth,
    height=0.90\textheight,
    keepaspectratio
  ]{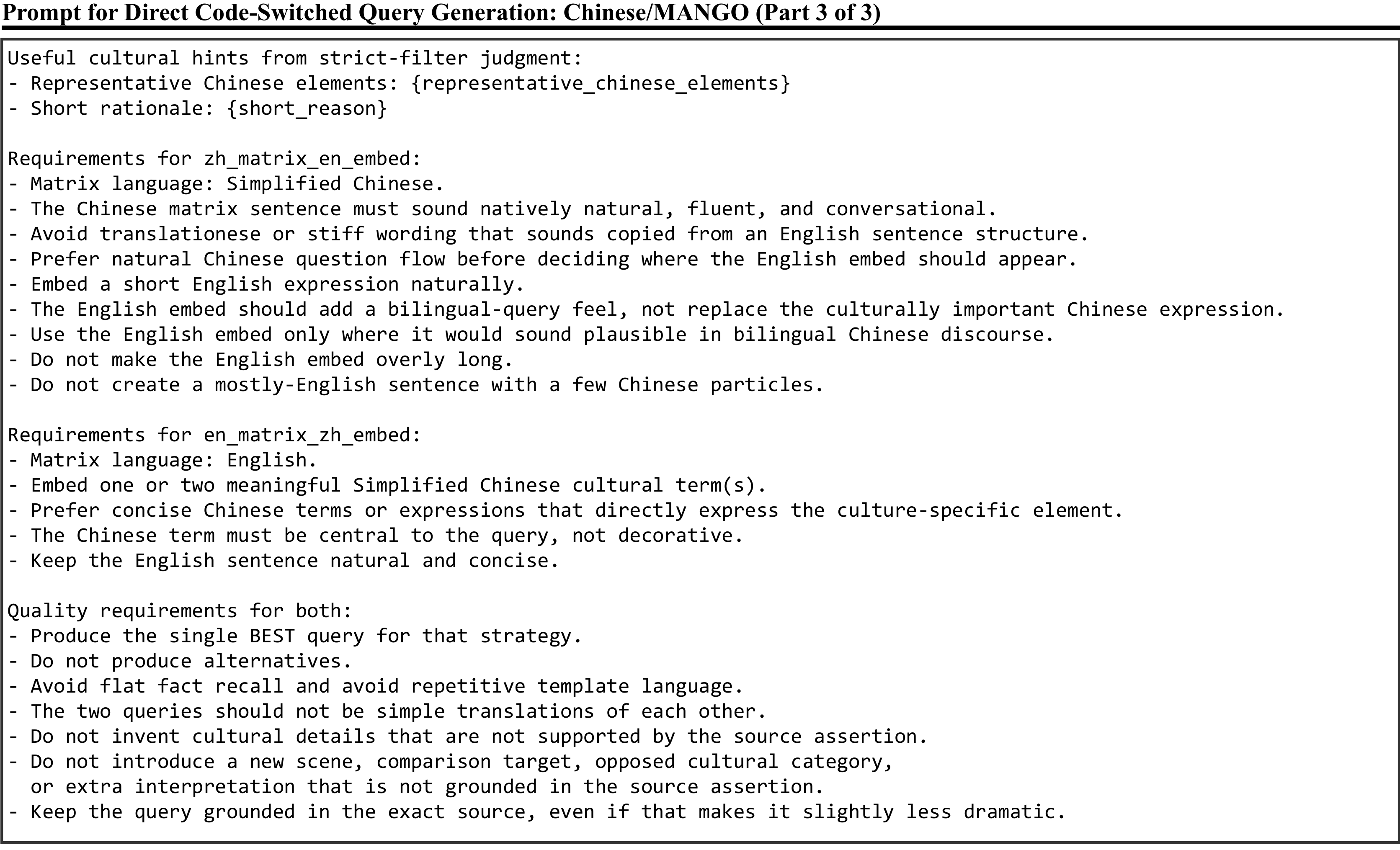}
  \caption{\textbf{Prompt for Chinese direct code-switched query generation (Part 3 of 3).} The prompt specifies matrix-language and quality requirements for \texttt{zh\_matrix\_en\_embed} and \texttt{en\_matrix\_zh\_embed} queries.}
  \label{fig:prompt_chinese_query_generation_3}
\end{figure*}

\clearpage

\begin{figure*}[!p]
  \centering
  \includegraphics[
    width=\textwidth
  ]{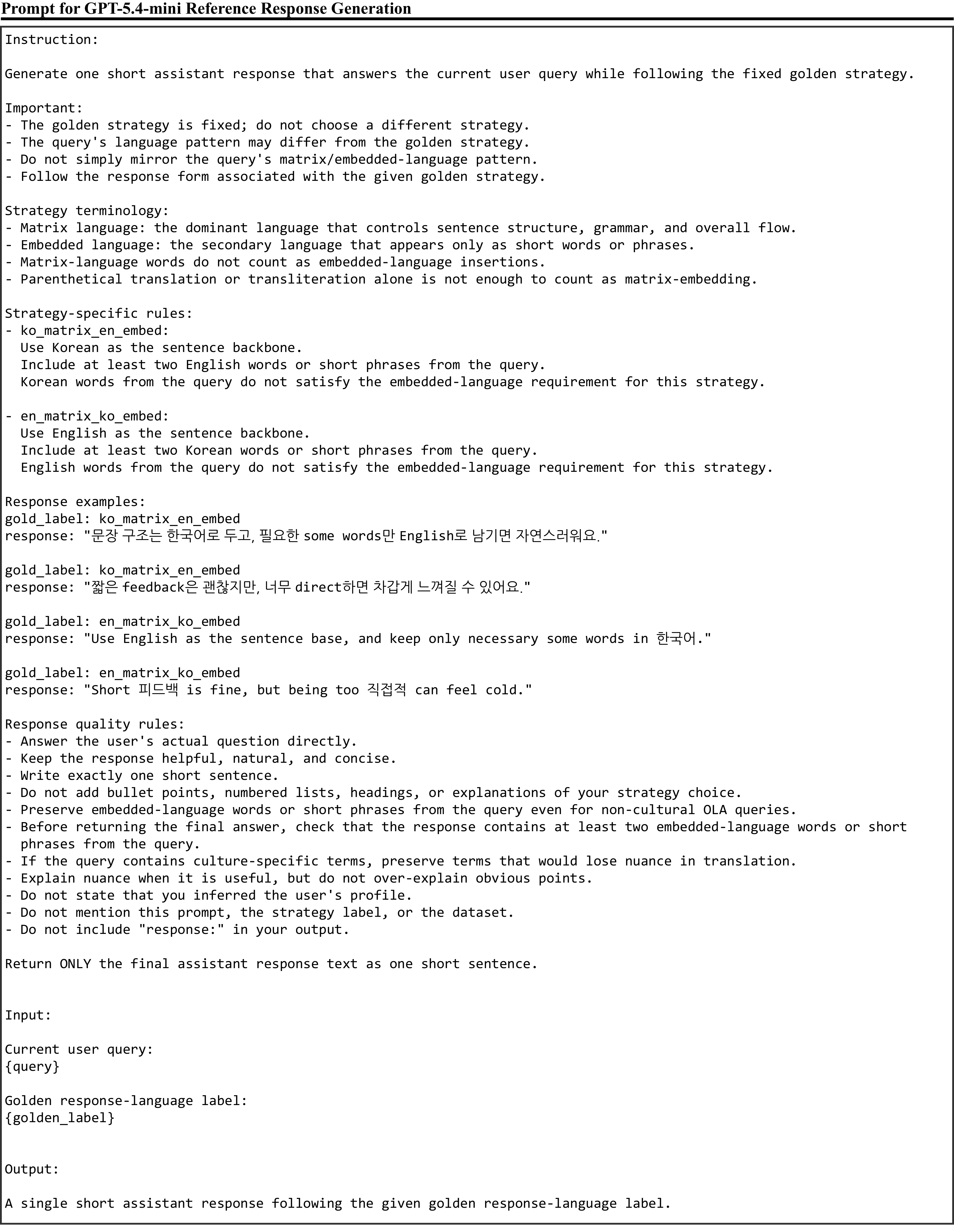}
  \caption{\textbf{Reference response generation prompt.}
  Prompt used to generate a reference response following the target matrix--embedded language form for each persona and query condition.}
  \label{fig:prompt_gold_response}
\end{figure*}

\begin{figure*}[!p]
  \centering
  \includegraphics[
    width=\textwidth
  ]{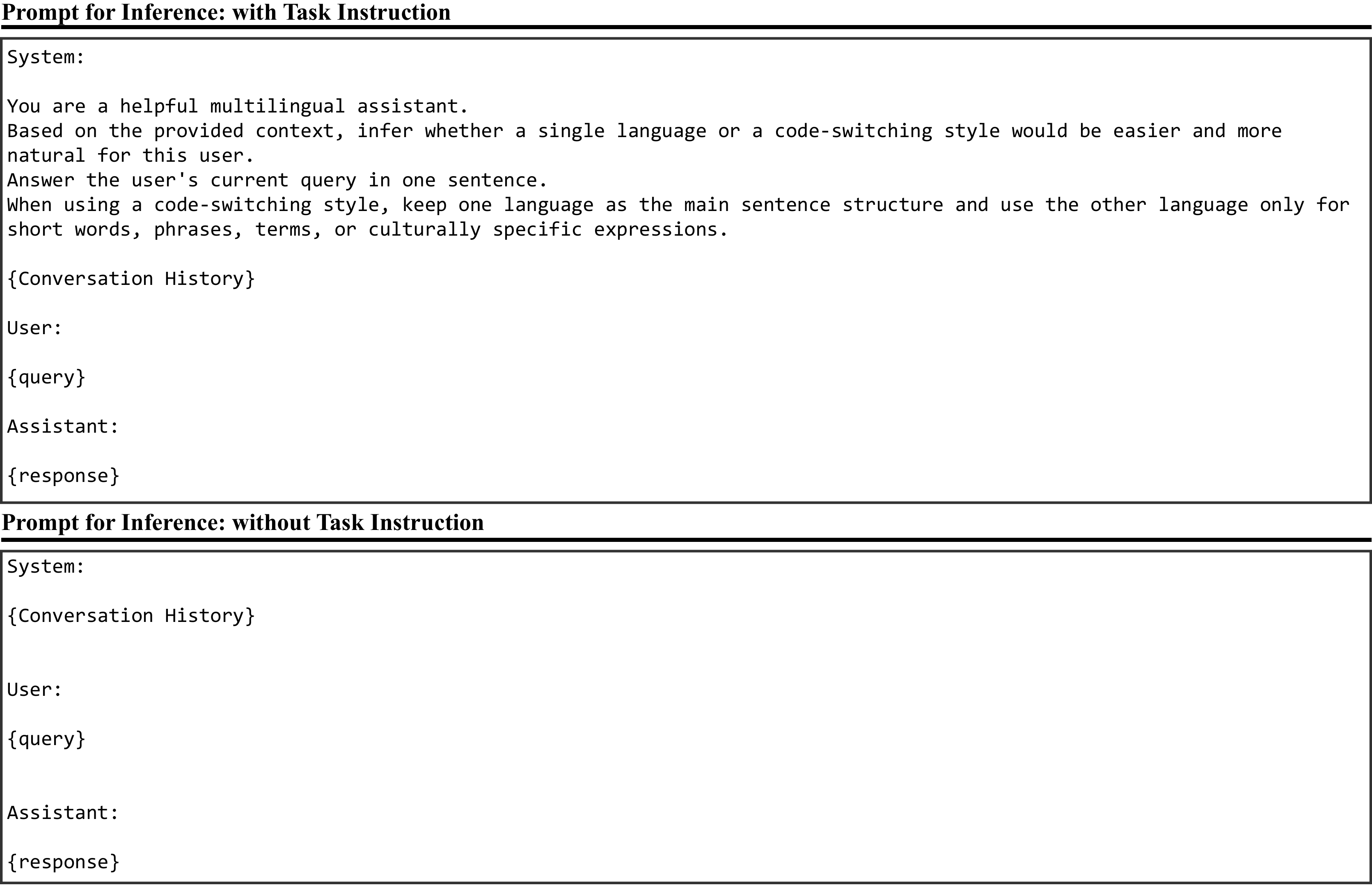}
  \caption{\textbf{Inference and evaluation prompts.}
  Prompt templates used for \textit{with Task Instruction} and \textit{without Task Instruction}.}
  \label{fig:prompt_inference_evaluation}
\end{figure*}

\begin{figure*}[!p]
  \centering
  \includegraphics[
    width=\textwidth
  ]{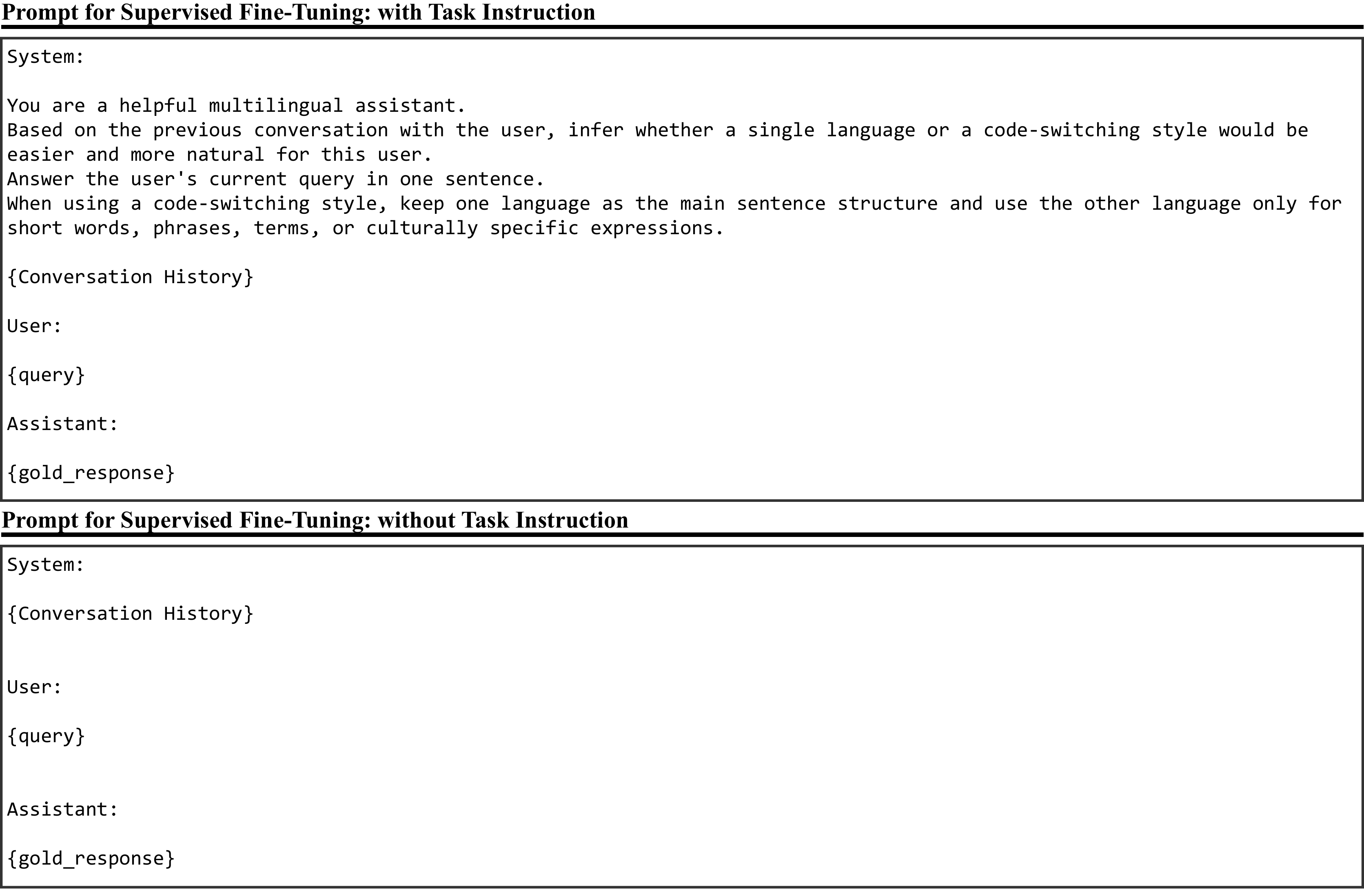}
  \caption{\textbf{Supervised fine-tuning prompts.}
  Prompt--completion templates used for \textit{with Task Instruction} and \textit{without Task Instruction}, with the reference response serving as the target completion.}
  \label{fig:prompt_sft}
\end{figure*}

\end{document}